\documentclass{article}

\usepackage{xcolor}

\usepackage[numbers,sort&compress]{natbib}
\usepackage[main, final]{neurips_2026}

\usepackage[utf8]{inputenc} 
\usepackage[T1]{fontenc}    
\usepackage{hyperref}       
\usepackage{url}            
\usepackage{booktabs}       
\usepackage{subcaption}
\usepackage{amsfonts}       
\usepackage{nicefrac}       
\usepackage{microtype}      
\usepackage{graphicx}
\graphicspath{{figs/}{./figs/}}
\usepackage{algorithm}
\usepackage{algpseudocode}

\definecolor{mine}{RGB}{205, 232, 248}
\definecolor{mydarkred}{RGB}{139, 0, 0}

\definecolor{rosered}{RGB}{255, 0, 127}
\definecolor{green}{RGB}{0, 128, 0}
\definecolor{orange}{RGB}{255, 165, 0}
\definecolor{darkblue}{RGB}{0, 0, 139}

\usepackage{multirow}
\usepackage{multicol}
\usepackage{varwidth} 

\usepackage{amsmath}
\usepackage{amssymb}
\usepackage{mathtools}
\usepackage{amsthm}
\usepackage{siunitx}

\usepackage{placeins}

\theoremstyle{plain}
\newtheorem{theorem}{Theorem}[section]

\theoremstyle{definition}

\theoremstyle{remark}

\usepackage{bm}
\newcommand{\bs}[1]{\boldsymbol{#1}}

\title{Towards Optimism-Pessimism Trade-off in Model-based Offline-to-Online Reinforcement Learning}

\begin{document}

\author{%
  \parbox{\textwidth}{%
    \centering
    \normalfont\normalsize
    \textbf{Guochen Zhou}$^{1,\dagger}$ \quad
    \textbf{Yijun Yang}$^{2,\dagger}$ \quad
    \textbf{Qiqi Duan}$^{3}$ \quad
    \textbf{Qing Su}$^{4}$ \quad
    \textbf{Weiming Ou}$^{5}$ \\[0.3em]
    \textbf{Li Shen}$^{6}$ \quad
    \textbf{Shihao Ji}$^{4}$ \quad
    \textbf{Yuhui Shi}$^{7,*}$ \quad
    \textbf{Zhisong Zhang}$^{1,*}$ \\[0.65em]
    {\normalfont
      $^{1}$Department of Computer Science, City University of Hong Kong \quad
      $^{2}$Tsinghua University \\[0.15em]
      $^{3}$Jiangxi University of Finance and Economics \quad $^{4}$University of Connecticut \\[0.15em]
      $^{5}$Shanghai University of Finance and Economics \quad
      $^{6}$Sun Yat-sen University \\[0.15em]
      $^{7}$Southern University of Science and Technology
    }%
  }%
}

\maketitle
\begingroup
  \renewcommand{\thefootnote}{\fnsymbol{footnote}}

  \footnotetext[1]{Corresponding authors.}

  \footnotetext[2]{Co-first authors.
  Contact email:
  \texttt{guochzhou2-c@my.cityu.edu.hk}.}
\endgroup

\begin{abstract}

Model-based offline-to-online reinforcement learning (RL) enables sample-efficient adaptation by leveraging offline pre-training for online fine-tuning. However, the distribution shifts between offline and online stages often hinder fine-tuning performance. Many existing methods encode the optimism-pessimism trade-off through a single-objective formulation. Selecting its trade-off coefficient through online fine-tuning can require costly bi-level optimization. We identify the balance established during offline training as a key challenge: in our motivating experiments, optimistic policies adapt better to novel tasks, whereas pessimistic policies perform better on the same online tasks. To address this challenge, we propose a bi-objective formulation that captures this trade-off, yielding a Pareto policy pool during offline training. This policy pool enables adaptive selection during online fine-tuning across diverse online tasks. To generate the pool, we introduce Multi-Objective Soft Actor-critIC (MOSAIC), which solves bi-objective problems and constructs diverse Pareto policies. After offline training, a contextual bandit algorithm hierarchically selects the most suitable policy for fine-tuning at each online interaction step. Empirically, our pipeline, \textbf{Hi}erarchical \textbf{P}areto \textbf{P}olicy \textbf{P}ool (\textbf{HiP3}), achieves the strongest aggregate performance across both same and novel online tasks and maintains its advantage under stronger distribution shifts. Comprehensive ablations further clarify the roles of the policy pool, MOSAIC-based pool construction, and the online selection mechanism. \looseness -1
\end{abstract}

\vspace{-5pt}
\section{Introduction}
\vspace{-2pt}
A major obstacle to deploying reinforcement learning (RL) in real-world settings is its high sample complexity. Offline RL mitigates this issue by learning from pre-collected data~\citep{levine2020offline}. However, distribution shifts between offline data and online interaction can hinder the deployment of offline-trained policies. This motivates offline-to-online RL (O2O RL), which adapts pretrained policies using limited online interaction~\citep{nakamoto2023cal,zhang2023policy}. Model-based O2O RL further improves sample efficiency by learning environment models for offline pre-training and online fine-tuning~\citep{mao2022moore,rafailov2023moto}.

\begin{figure*}
	\centering
    \includegraphics[width=0.85\linewidth]{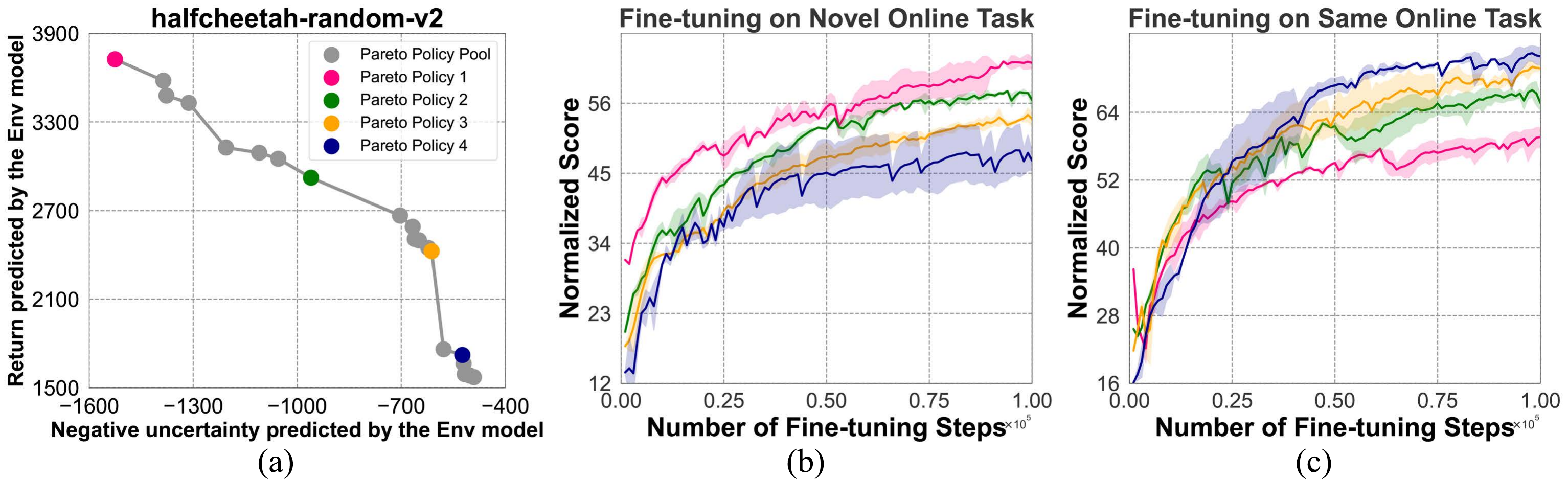}
    \vspace{-9pt}
        \caption{\footnotesize (a) Pareto policy pool discovered by HiP3, containing policies with different optimism-pessimism trade-offs. \textcolor{rosered}{P1} maximizes model-predicted return (most optimistic), \textcolor{darkblue}{P4} minimizes uncertainty (most pessimistic), while \textcolor{green}{P2} and \textcolor{orange}{P3} represent balanced trade-offs. (b) On the novel online task (details in Appendix~\ref{app:environments}), optimistic \textcolor{rosered}{P1} achieves the highest return. (c) On the same online task, pessimistic \textcolor{darkblue}{P4} achieves the highest return. Together, these results motivate preserving multiple trade-off policies offline and deferring the choice among them to online adaptation, rather than committing to a single trade-off in advance.} \vspace{-1mm}
    \label{fig:pareto_front}
    \vspace{-1mm}
\end{figure*}

Despite this progress, distribution shifts continue to limit fine-tuning performance in model-based O2O RL. We argue that a key factor is the optimism-pessimism trade-off established during offline training. Following prior works~\citep{yu2020mopo,kidambi2020morel,yang2021pareto}, we use model-predicted return and uncertainty as proxies for optimism and pessimism, respectively. Figure~\ref{fig:pareto_front}(a) illustrates offline policies with different trade-offs between these objectives. On the novel online task with controlled dynamics and reward shifts, the optimistic policy \textcolor{rosered}{P1} (higher model return) performs best in Figure~\ref{fig:pareto_front}(b). Conversely, the pessimistic policy \textcolor{darkblue}{P4} (lower model uncertainty) performs best on the same online task in Figure~\ref{fig:pareto_front}(c). This reversal is consistent with an optimistic bias that may support exploration beyond the offline data distribution and a pessimistic bias toward high-confidence regions. Policies \textcolor{green}{P2} and \textcolor{orange}{P3} represent intermediate trade-offs. Appendix~\ref{app:motivation} further reproduces this reversal across four controlled settings spanning different tasks and offline-data qualities. These observations suggest that the preferred optimism-pessimism balance depends on the task encountered during online fine-tuning.

Existing model-based approaches commonly encode this trade-off through a single scalarized objective, such as $\hat r-\lambda u$, thereby committing each offline policy to a particular balance. Adapting this balance to an online task by selecting $\lambda$ according to fine-tuning performance requires an outer search over coefficients around the inner policy-optimization procedure, making task-specific tuning costly. Training multiple policies with different fixed coefficients is a natural alternative, but linear scalarization does not in general recover all Pareto-optimal trade-offs when the attainable objective set is non-convex. Appendix~\ref{app:ls_comparison} provides a detailed discussion and empirical comparison.

In this paper, we treat optimism and pessimism as two distinct objectives and introduce Hierarchical Pareto Policy Pool (HiP3), which constructs a diverse offline policy pool for adaptive online selection. During online adaptation, the agent encounters changing states and updates its candidate policies, making policy selection an ongoing decision. This design raises two algorithmic challenges. First, how can we generate policies that cover diverse optimism-pessimism trade-offs across the Pareto front? Second, how can we dynamically select a suitable policy for the current state during online fine-tuning, rather than committing to a single candidate throughout adaptation? To address the first challenge, we introduce Multi-Objective Soft Actor-critIC (MOSAIC), which solves constrained bi-objective policy optimization problems defined by reference vectors targeting different regions of the Pareto front. It then uses warm-started neighborhood search to discover additional trade-off policies around the obtained ones. To address the second challenge, HiP3 uses a contextual bandit as a high-level controller over a candidate subset of the offline policy pool. At each online interaction step, the controller selects a policy conditioned on the current state, and the selected policy is fine-tuned with model-based updates.

We evaluate HiP3 on the standard D4RL continuous-control benchmark~\citep{fu2020d4rl} against representative online and O2O RL baselines across both same and novel online tasks. HiP3 achieves the strongest average performance in both settings, with particularly clear advantages on novel tasks, and remains effective under stronger distribution shifts. Ablation studies examine the roles of the policy pool, MOSAIC-based pool construction, state-dependent policy selection, and candidate-set initialization. Additional analyses characterize policy-selection behavior during online adaptation. Our code is available at \url{https://github.com/GuochenZhou/HiP3}

\vspace{-5pt}
\section{Related Work} 
\vspace{-2pt}
\textbf{Offline RL and Model-based Variants.} Offline RL trains policies on static datasets to avoid the high sample complexity of online methods~\citep{levine2020offline}. A major challenge is poor extrapolation to out-of-distribution (OOD) actions, often causing overestimation. Most approaches rely on pessimism to counter this, via policy constraints~\citep{fujimoto2019off,kumar2020conservative}, regularization~\citep{wang2020critic}, or uncertainty penalties~\citep{wu2019behavior,kumar2019stabilizing}. Model-based methods improve efficiency with learned dynamics for planning or simulated rollouts~\citep{janner2019trust,argenson2020model,matsushima2021deployment,yu2021combo}. Notably, works like MOPO~\citep{yu2020mopo} and MOReL~\citep{kidambi2020morel} incorporate uncertainty to penalize OOD regions (e.g., subtracting it from rewards). Critically, these fix the optimism-pessimism trade-off via a single penalty, limiting flexibility for online fine-tuning.

\textbf{Offline-to-Online (O2O) RL.} Distribution shifts hinder direct online deployment of offline policies~\citep{koh2021wilds}, necessitating fine-tuning. However, fine-tuning often underperforms~\citep{nair2020awac} due to (1) training instability from the shift~\citep{lee2022offline}, and (2) restricted exploration inherited from offline pessimism. Instability is mitigated via balanced replay~\citep{lee2022offline}, priority sampling~\citep{zheng2023adaptive}, policy expansion~\citep{zhang2023policy}, state-adaptive policy families~\citep{wang2023train}, and more recently warm-start recalibration without retaining offline data~\citep{zhouefficient2025}. Exploration or policy improvement is enhanced via critic guidance~\citep{mark2022fine}, guide policies~\citep{uchendu2023jump}, perturbed value updates and more frequent critic updates~\citep{zhang2024perspective}, optimistic critic reconstruction with constrained fine-tuning~\citep{luo2024optimistic}, energy-guided diffusion sampling that distills offline knowledge for online data generation~\citep{liuenergy2024}, and expressive flow-based policies~\citep{fql_park2025}. Beyond O2O-specific methods, scalable world-model agents such as TD-MPC2~\citep{hansen2024tdmpc2} provide strong sample-efficient model-based baselines, but do not explicitly address the optimism-pessimism trade-off induced by offline pretraining. Most of these methods optimize a single policy or otherwise commit to a fixed adaptation strategy, rather than preserving multiple optimism-pessimism trade-offs for later selection. HiP3 instead treats the trade-off itself as an adaptation decision: it preserves diverse trade-off policies offline and performs state-dependent policy selection among them during online adaptation.

\vspace{-5pt}
\section{Preliminaries}\vspace{-2pt}
Standard RL formulates agent-environment interaction as a Markov Decision Process (MDP) $(S, A, p, r, \rho_0, \gamma)$~\citep{sutton2018reinforcement}, where the goal is to find a policy $\pi_\theta$ maximizing expected discounted return:
\vspace{-4pt}
\begin{equation}
    J(\pi_\theta)= \mathbb{E}_{\rho_0,\pi_{\theta}} \left [\sum_{t=0}^{+\infty}\gamma^t r(s_t,a_t) \right ].
    \label{eq:3_1}
    \vspace{-1pt}
\end{equation}
Specifically, in model-based O2O RL, the agent first learns an environment model and a policy from pre-collected data, then fine-tunes them using limited online interaction~\citep{mao2022moore, rafailov2023moto}. The environment model, typically parameterized by a feed-forward neural network, contains a learned transition probability model $\hat{p}(s'|s, a)$ and a reward model $\hat{r}(s, a)$. Given the current state $s$ and action $a$ selected by policy $\pi_{\theta}(s)$, it predicts the next state $s'\sim \hat{p}(s'|s, a)$ and reward value $\hat{r}$.\looseness-1

These model predictions naturally expose the optimism-pessimism trade-off studied in this work. Following prior works~\citep{yu2020mopo,kidambi2020morel,yang2021pareto}, we use the model-predicted reward $\hat{r}(s, a)$ and uncertainty estimate $u(s, a)$ as proxies for the optimism and pessimism objectives, respectively~\footnote{Further information on the uncertainty estimation $u$ is provided in Appendix~\ref{appendix: uncertainty estimation}, and a description of the environment model architecture is given in Appendix~\ref{appendix: environmental model}.}. A common formulation combines the two terms through linear scalarization, e.g., $\overline{r}(s, a) = \hat{r}(s, a)-\lambda u(s, a)$, where $\lambda$ specifies a global optimism-pessimism balance. For a new online task, selecting an appropriate $\lambda$ through target-task rollouts and online fine-tuning requires an outer search over $\lambda$ around the inner policy-optimization procedure, leading to costly bi-level tuning.

Rather than selecting a single trade-off coefficient online, we formulate optimism and pessimism as two distinct objectives and solve the following bi-objective optimization problem. Appendix~\ref{app:ls_comparison} provides a detailed comparison with linear scalarization.
\vspace{-4pt}
\begin{equation}
\label{eq:3_2}
\begin{aligned}
  \max_{\theta} J(\pi_{\theta}) &= \max_{\theta} \bigl( J^{\hat{r}}(\pi_\theta),\; J^{\hat{u}}(\pi_\theta) \bigr)
\end{aligned}
\vspace{-1pt}
\end{equation}
where $\hat{u}(s_t,a_t)\triangleq\exp(-u(s_t,a_t)/\kappa)$ maps the raw uncertainty to a bounded score in $(0,1]$, so that larger uncertainty yields a smaller objective value. We use $\kappa=1.5$ by default, except when varying it in the sensitivity analysis. This bi-objective formulation replaces a single scalarized balance with explicit optimism and pessimism objectives, enabling offline construction of policies spanning different trade-off regions for subsequent state-dependent policy selection. Appendix~\ref{app:moo} provides a brief introduction to multi-objective optimization.

HiP3 instantiates this formulation with MOSAIC for offline policy-pool construction and a contextual bandit for state-dependent policy selection during online adaptation.

\vspace{-5pt}
\section{HiP3: Hierarchical Pareto Policy Pool}\label{sec: method}
\vspace{-2pt}
\begin{figure*}[t]
    \centering
    \includegraphics[width=1.0\linewidth]{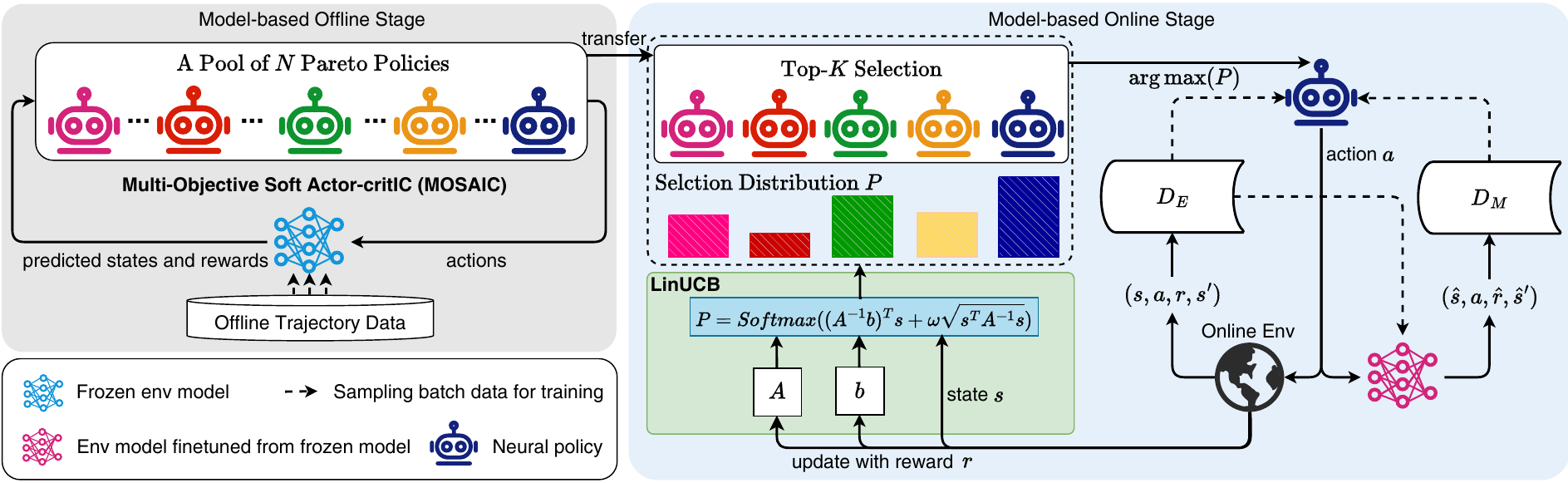}
    \vspace{-15pt}
    \caption{\footnotesize
    Overview of Hierarchical Pareto Policy Pool (HiP3). In the offline stage (gray box), our MOSAIC algorithm generates a diverse pool of Pareto policies exhibiting varying optimism-pessimism trade-offs via multi-objective optimization with reference vectors and neighborhood search. In the online stage (blue box), a contextual bandit (LinUCB~\citep{chu2011contextual}) selects a candidate policy conditioned on the current state from the top-$K$ subset; the selected policy is then fine-tuned via a Dyna-style method~\citep{janner2019trust}.}
    \label{fig:HiP3}
    \vspace{-2mm}
\end{figure*}


Figure~\ref{fig:HiP3} provides an overview of HiP3 alongside the procedure summarized in Algorithm~\ref{algo:HiP3}. Rather than committing offline training to a single optimism-pessimism balance, HiP3 preserves multiple trade-offs offline and performs state-dependent policy selection during online adaptation. In the offline stage, MOSAIC constructs a finite Pareto policy pool spanning different optimism-pessimism trade-offs through reference-vector-guided multi-objective optimization and neighborhood search. In the online stage, HiP3 first uses one-time ranking to select a candidate subset from the policy pool and then uses a contextual bandit to select a candidate policy conditioned on the current state. The selected policy is then fine-tuned online.

\begin{algorithm}[htb]
\caption{Hierarchical Pareto Policy Pool (HiP3)}
\label{algo:HiP3}
\small

\noindent \textbf{Input:} Offline dataset $\mathcal{D}$, \#reference vectors $n$, \#offline policies $K_\textrm{off}$, \#online policies $K_\textrm{on}$, \#offline steps $T_\textrm{off}\!=\!n(T_g\!+\!2T_l)T_e$, \#online steps $T_\textrm{on}$, model update interval $S_M$, bandit hyperparameter $\omega$ \\
\noindent \textbf{Initialize:} Pareto policy pool $\mathcal{P} = \emptyset$, env. model $M$, model data buffer $\mathcal{D}_M = \emptyset$, env. data buffer $\mathcal{D}_E = \emptyset$

\setlength{\columnsep}{5pt} 
\vspace{-8pt} 

\begin{multicols}{2}
\begin{algorithmic}[1] 
\setlength{\leftmargin}{1em}
\algnotext{EndFor}
\algnotext{EndIf}

\Statex \textcolor{blue}{// Offline learning via MOSAIC}
\State Train $M$ on dataset $\mathcal{D}$ (max. likelihood)
\State Generate reference vectors $\{\bs{v}_1,\dots,\bs{v}_n\}$ (Eq.~\ref{eq:4_1})
\For{$i = 1$ to $n$} {\scriptsize//with $\mathcal{D}_M$ accumulated}
    \State Initialize a policy $\pi_\theta$
    \State Update $\pi_\theta$ for $T_g\!\times\!T_e$ steps with $\bs{v}_i$ (Alg.~\ref{algo:mo_sac})
    \State $\bs{v}_i^+ \gets \bs{v}_i + \epsilon$; $\bs{v}_i^- \gets \bs{v}_i - \epsilon$
    \For{$\bs{v}' \in \{\bs{v}_i^+, \bs{v}_i^-\}$}
        \For{$j = 1$ to $T_l$}
            \State  Update $\pi_\theta$ for $T_e$ steps with $\bs{v}'$ (Alg.~\ref{algo:mo_sac})
            \State $\mathcal{P} \gets \mathcal{P} \cup \{\pi_\theta\}$
            \State $\mathcal{P}\gets \text{top-}K_\textrm{off}(\mathcal{P})$ {\scriptsize//non-dom. sort}
        \EndFor
    \EndFor
\EndFor

\columnbreak 

\Statex \textcolor{blue}{// State-dependent policy selection via LinUCB}
\State $\mathcal{D}_M = \emptyset$ 
\State $\mathcal P_{\rm on} \leftarrow$ top-$K_{\rm on}$ policies by rollout ranking
\State Construct contextual bandit $B(A, b)$ with $\mathcal{P}_{on}$
\For{$t = 1$ to $T_\textrm{on}$}
    \If{$t \pmod{S_M} == 0$}
        \State Train $M$ on $\mathcal{D}_E$ (max. likelihood)
        \State $\mathcal{D}_M \gets \mathcal{D}_M \cup \{\text{rollout}(M, B, \mathcal{P}_{on})\}$
    \EndIf
    \State $P_t \leftarrow \mathrm{Softmax}\!\left((A^{-1}b)^\top s_t +\omega\sqrt{s_t^\top A^{-1}s_t}\right)$
    \State $k_t \leftarrow \arg\max_k P_t(k)$
    \State Execute $\pi^{k_t}_\theta \in \mathcal{P}_{on}$ in env. $E$, observe $r, \tau$
    \State $A_{k_t} \gets A_{k_t} + \eta_B s_t s_t^T$; $b_{k_t} \gets b_{k_t} + \eta_B s_t r_t$
    \State $ \mathcal{D}_E \gets  \mathcal{D}_E \cup \{\tau\}$
    \State Update $\pi^{k_t}_\theta$ via SAC on $\mathcal{D}_E \cup \mathcal{D}_M$
\EndFor
\end{algorithmic}
\end{multicols}
\vspace{-10pt}
\end{algorithm}

\vspace{-2mm}
\subsection{MOSAIC: Generation of Pareto Policy Pool}\vspace{-1mm}

The bi-objective policy optimization problem in Eq.~\ref{eq:3_2} may admit many Pareto-optimal policies, making it infeasible in practice to enumerate the entire Pareto set. Reference vectors provide an explicit way to target different regions of the bi-objective space without prescribing a single global scalarization. Following prior works~\citep{lin2019pareto, mahapatra2020multi, chen2022multi}, we instantiate these regions using $n$ evenly spaced reference vectors $\{\bs{v}_i\}_{i=1}^{n}$, each specifying a target trade-off between the two objectives. Specifically, we construct the reference vectors as follows:

\vspace{-2pt}
\begin{equation}
\begin{aligned}
    &\bs{v}_i=(v_i^1,v_i^2) = \left(\tau_b-(i-1) \tau_c, \; \tau_a+(i-1) \tau_c\right),
    &\text{with } \tau_c=\frac{\tau_b-\tau_a}{n-1}, \;\; 0<\tau_a<\tau_b<1, \;\; i=1,\cdots,n
\end{aligned}
\label{eq:4_1}
\end{equation}
where $\tau_a$ and $\tau_b$ control the range covered by the reference vectors. Each $\bs{v}_i=(v_i^1,v_i^2)$ defines a constrained bi-objective optimization problem whose solution lies in a targeted region of the Pareto front: 
\vspace{-7pt}
\begin{equation}
\begin{aligned}    
    &\max_{\theta}\bs{J}(\pi_{\theta}) = \max_{\theta}\Bigl(J^{\hat{r}}(\pi_\theta),J^{\hat{u}}(\pi_\theta)\Bigr)\\
    &\textrm{s.t.}\; \Psi(\pi_{\theta}, \bs{v}_i)=D_{\cos}\Bigl(\frac{\bs{J}(\pi_\theta)}{\|\bs{J}(\pi_\theta)\|_2},\frac{\bs{v}_i}{\|\bs{v}_i\|_2}\Bigr)\geq{\psi},
    \label{eq:4_2}
\end{aligned}
\end{equation}
where $\Psi(\pi_{\theta}, \bs{v}_i)$ defines the cosine similarity between the objective vector $\bs{J}(\pi_{\theta})$ and the reference vector $\bs{v}_i$, and the hyper-parameter $\psi$ controls the scope of the target region. To solve this bi-objective optimization problem with an inequality constraint, we introduce MOSAIC, a multi-objective extension of SAC, with the pseudocode provided in Algorithm~\ref{algo:mo_sac}. As an off-policy SAC-style algorithm, MOSAIC learns two soft critics associated with $\hat{r}$ and $\hat{u}$. Their Bellman targets follow the entropy-regularized form:
\begin{equation}
\begin{aligned}
    Q^{\hat{r}}_{\phi^{\hat{r}}}(s_t, a_t)
    &= \hat{r}(s_t, a_t)
    + \gamma \mathbb{E}_{s_{t+1},\, a_{t+1}\sim \pi_{\theta}}
    \left[
        Q^{\hat{r}}_{\overline{\phi}^{\hat{r}}}(s_{t+1}, a_{t+1})
        - \alpha \log \pi_{\theta}(a_{t+1}\mid s_{t+1})
    \right], \\
    Q^{\hat{u}}_{\phi^{\hat{u}}}(s_t, a_t)
    &= \hat{u}(s_t, a_t)
    + \gamma \mathbb{E}_{s_{t+1},\, a_{t+1}\sim \pi_{\theta}}
    \left[
        Q^{\hat{u}}_{\overline{\phi}^{\hat{u}}}(s_{t+1}, a_{t+1})
        - \alpha \log \pi_{\theta}(a_{t+1}\mid s_{t+1})
    \right].
\end{aligned}
\label{eq:q_function}
\end{equation}

During each update iteration, the two soft critics and their target networks are updated by minimizing the corresponding entropy-regularized Bellman residuals on batches sampled from the model buffer $\mathcal{D}_M$, as outlined in Lines 6-9 of Algorithm~\ref{algo:mo_sac}. MOSAIC then checks whether the current policy lies in the target region induced by the reference vector. If not, the {\it Correction stage} prioritizes the objective that is currently under-emphasized relative to the trade-off prescribed by $\bs{v}_i$. Once the alignment criterion is met, MOSAIC enters the {\it Ascent stage} and treats the alignment term as the third objective:
\vspace{-3pt}
\begin{equation}
    \max_\theta\bs{F}(\pi_\theta) \triangleq \max_\theta \left(J^{\hat{r}}(\pi_{\theta}), J^{\hat{u}}(\pi_{\theta}), \Psi(\pi_{\theta}, \bs{v}_i)\right).
    \label{eq:4_4}
\end{equation}

In Algorithm~\ref{algo:mo_sac}, ${L^{Q^{\hat{r}}}}$, ${L^{Q^{\hat{u}}}}$ and $L^{\alpha}$ denote the standard SAC soft-critic losses for $\hat{r}$ and $\hat{u}$, and the temperature loss, respectively. In practice, $J^{\hat{r}}$ and $J^{\hat{u}}$ denote the SAC-style actor surrogates induced by the two critics, and the ascent stage combines $\nabla_{\theta} J^{\hat{r}}(\pi_{\theta})$, $\nabla_{\theta} J^{\hat{u}}(\pi_{\theta})$ and $\nabla_{\theta}\Psi$ through MGDA. For analysis, we use the equivalent objective-space view, where minimizing the actor-side surrogates corresponds to ascending the paired objectives. We then solve the resulting tri-objective problem with MGDA~\citep{desideri2012multiple}, which finds coefficients $\beta_t$ for a convex combination of the three objective gradients, yielding a common ascent direction that does not decrease any objective. This is equivalent to the following min-norm problem:
\vspace{-2pt}
\begin{equation}	
	\min_{\bs{\beta}_t} \|\bs{\beta}_t \nabla_{\theta_t}\bs{F}(\pi_{\theta_t})\|_2 \,\,\,\, s.t.\,\,\,\, \|\bs{\beta}_t\|_1=1,\bs{\beta}_t\geq0,
	\label{eq:4_5}
\end{equation}

which can be effectively tackled via the Frank-Wolfe algorithm~\citep{jaggi2013revisiting}. Subsequently, we perform the policy update as specified in Line 17 of Algorithm~\ref{algo:mo_sac}. The resulting ascent stage optimizes the three-objective formulation above and admits a first-order convergence guarantee toward Pareto stationarity in expectation. A complete proof is provided in Appendix~\ref{appendix: convergence analysis}, where Theorem~\ref{theorem1} establishes the convergence rate of the ascent stage of Algorithm~\ref{algo:mo_sac}. For the correction stage, which reduces algorithmically to a standard SAC-style single-objective surrogate update, we focus the new analysis on the ascent stage and refer to prior SAC~\citep{haarnoja2018soft} work for the underlying update form.

MOSAIC uses multiple reference vectors to generate Pareto policies with diverse optimism-pessimism trade-offs. However, using too few reference vectors may miss useful trade-offs and lead to suboptimal policy selection in the online stage. To address this problem, MOSAIC expands trade-off coverage through warm-started neighborhood search.

For each reference vector $v_i$, we perturb it to create augmented reference vectors $v_i^+$ and $v_i^-$ in Line 6 of Algorithm~\ref{algo:HiP3}, thereby constructing additional constrained bi-objective optimization problems based on Eq.~\ref{eq:4_2}. By solving these bi-objective problems in the neighborhood of the original ones, additional trade-off candidates can be obtained. The resulting candidates are added to the policy pool, followed by non-dominated sorting. This procedure reuses previously optimized policies to populate additional trade-off regions within the allocated offline budget.

\begin{algorithm}[t]
\caption{Multi-Objective Soft Actor-critIC (MOSAIC)}
\vspace{-15pt}
\label{algo:mo_sac}
\begin{multicols}{2}
\small
\algnotext{EndFor}
\algnotext{EndIf}
\small
\noindent \textbf{Input:} ref. vector $\bs{v}$, threshold $\psi$, \#iterations $T$, env. model $M$, data buffer $\mathcal{D}_M$ and its update interval $S_M$ \\
\noindent \textbf{Initialize:} $\phi^{\hat{r}}_i$, $\phi^{\hat{u}}_i$, $\overline{\phi_i^{\hat{r}}}$, $\overline{\phi_i^{\hat{u}}}$ for $i\in\{1,2\}$

\begin{algorithmic}[1]
\For{$t=1$ to $T$}
    \If{$t\pmod{S_M} == 0$}
        \State Obtain $\tau_M$ with $k$-step rollout on $M$ by $\pi_{\theta}$
        \State $\mathcal{D}_M \gets \mathcal{D}_M \cup \{\tau_M\}$
    \EndIf
    \State Sample a batch from $\mathcal{D}_M$
    \State $\phi^{\hat{r}}_{i} \gets \phi^{\hat{r}}_{i} - \eta \nabla_{\phi^{\hat{r}}_i} L^{Q^{\hat{r}}}(\phi^{\hat{r}}_i)$ for $i \in \{1,2\}$
    \State $\phi^{\hat{u}}_i \gets \phi^{\hat{u}}_i - \eta \nabla_{\phi^{\hat{u}}_i} L^{Q^{\hat{u}}}(\phi^{\hat{u}}_i)$ for $i \in \{1,2\}$
    \State $\overline{\phi^{\hat{r}}_i} \gets \tau \phi^{\hat{r}}_i + (1 - \tau) \overline{\phi^{\hat{r}}_i}$ for $i \in \{1,2\}$
    \State $\overline{\phi^{\hat{u}}_i} \gets \tau \phi^{\hat{u}}_i + (1 - \tau) \overline{\phi^{\hat{u}}_i}$ for $i \in \{1,2\}$
\Statex 
\Statex \textcolor{blue}{// Correction stage}
\If{$\Psi(\pi_{\theta}, \bs{v}) < \psi$}
    \If{$\frac{J^{\hat{r}}(\pi_{\theta})}{J^{\hat{u}}(\pi_{\theta})} < \frac{\bs{v}^1}{\bs{v}^2}$}
        \State $\pi_{\theta} \gets \pi_{\theta} + \eta \nabla_{\theta} J^{\hat{r}}(\pi_{\theta})$
    \Else
        \State $\pi_{\theta} \gets \pi_{\theta} + \eta \nabla_{\theta} J^{\hat{u}}(\pi_{\theta})$
    \EndIf

\Statex \textcolor{blue}{// Ascent stage}
\Else
    \State Find $\bs{\beta}^*$ according to Eq.~\ref{eq:4_5}
    \State $\pi_{\theta} \gets \pi_{\theta} + \eta \bs{\beta}^* \nabla_{\theta} \bs{F}(\pi_{\theta})$
\EndIf
\State $\alpha \gets \alpha - \lambda \nabla_{\alpha} L^{\alpha}(\alpha)$ {\scriptsize//SAC temperature $\alpha$ update}
\EndFor

\Statex \textbf{Output:} policy $\pi_{\theta}$
\end{algorithmic}
\end{multicols}
\vspace{-10pt}
\end{algorithm}

\vspace{-2mm}
\subsection{State-Dependent Policy Selection and Online Optimization}
\vspace{-1mm}

The offline policy pool preserves multiple optimism-pessimism trade-offs, but online adaptation still requires deciding how to use this diversity. Committing to a single policy for the entire trajectory discards the possibility that different states favor different candidates. To use this offline diversity during online adaptation, HiP3 first narrows the pool to a compact candidate set and then performs state-dependent policy selection within this set.


Before fine-tuning, HiP3 performs a one-time rollout ranking on the target task and selects the top-$K$ policies as the online candidate set $\mathcal{P}_{on}$, without reusing the ranking trajectories for policy or model updates. A contextual bandit then serves as the high-level controller over these low-level candidate policies throughout online adaptation (Lines 12--24 of Algorithm~\ref{algo:HiP3} and the blue box in Figure~\ref{fig:HiP3}).

At each online interaction step $t$, LinUCB computes a confidence-adjusted score for each candidate policy from the current state $s_t$. For notational compactness, we write $\mathcal{P}_{on}=\{\pi_\theta^k\}_{k=1}^{K_{on}}$ and the LinUCB statistics as $B(A,b)$, where $A=\{A_k\}_{k=1}^{K_{on}}$ and $b=\{b_k\}_{k=1}^{K_{on}}$. We softmax-normalize the LinUCB scores into a policy-preference vector $P_t$ for visualization. Since softmax preserves their ordering, execution is equivalent to selecting the candidate with the highest raw LinUCB score. LinUCB then uses the observed environment reward as direct target-task feedback to update only the statistics $(A_{k_t},b_{k_t})$ of the executed policy. Additional details are provided in Appendix~\ref{app:bandit}.

After selection, only the executed policy is fine-tuned using a Dyna-style SAC update with real online transitions in $\mathcal{D}_E$ and short model-generated rollouts in $\mathcal{D}_M$. The original offline dataset $\mathcal{D}$ is not retained during online fine-tuning.\looseness-1

\vspace{-5pt}
\section{Experimental Results}
\vspace{-2pt}
\label{sec: experimental evaluation}
\subsection{Experimental Setup}
\vspace{-0.5mm}
We evaluate HiP3 against O2O RL and online RL baselines on two task categories derived from the D4RL benchmark~\citep{fu2020d4rl}: same online tasks and novel online tasks (see Appendix~\ref{app:environments} for a brief overview). Our primary evaluation target is adaptation under changes in dynamics and rewards. Fine-tuning on the same online tasks provides a complementary evaluation of the same framework.

We compare HiP3 with representative O2O RL baselines, including Cal-QL~\citep{nakamoto2023cal}, PEX~\citep{zhang2023policy}, AWAC~\citep{nair2020awac}, IQL~\citep{kostrikov2021offline}, CQL~\citep{kumar2020conservative}, and FamO2O~\citep{wang2023train}, together with the online RL baseline MBPO~\citep{janner2019trust}. CQL, Cal-QL, AWAC, and IQL are run using the publicly available CORL~\citep{tarasov2023corl} codebase with default hyperparameters unless otherwise specified, while PEX and MBPO use their official implementations. After 1M offline steps, all O2O methods except FamO2O are fine-tuned for 100K online steps, and MBPO is trained from scratch for 100K steps. For HiP3, the online candidate set is initialized by the one-time ranking described in Section~\ref{sec: method} before the 100K-step fine-tuning stage. Implementation details are provided in Appendix~\ref{appendix: baseline implementation}.

The offline policy-optimization budget is distributed across the complete policy pool rather than assigned independently to each policy, while each online update optimizes only the selected policy. Appendix~\ref{app:comput_cost} reports matched end-to-end computational costs together with additional stage-wise runtime measurements.

\vspace{-2mm}
\subsection{Main Results}
\vspace{-0.5mm}

\setlength{\tabcolsep}{2pt}
\begin{table*}[t]
\scriptsize
\centering
\footnotesize 
\setlength{\tabcolsep}{3pt} 

\resizebox{\textwidth}{!}{

\begin{tabular}
{c@{\hspace{5pt}}lr@{\hspace{-1pt}}lr@{\hspace{-1pt}}lr@{\hspace{-1pt}}lr@{\hspace{-1pt}}lr@{\hspace{-1pt}}l | r@{\hspace{-1pt}}lr@{\hspace{-1pt}}lr@{\hspace{-1pt}}lr@{\hspace{-1pt}}lr@{\hspace{-1pt}}l}
\toprule
 & & \multicolumn{10}{c}{\textbf{On the Same Online Tasks}} 
   & \multicolumn{10}{c}{\textbf{On Novel Online Tasks}} \\
\cmidrule(lr){3-12}\cmidrule(lr){13-22}
 &  & \multicolumn{2}{c}{AWAC} & \multicolumn{2}{c}{PEX} & \multicolumn{2}{c}{Cal-QL} &  \multicolumn{2}{c}{MBPO} & \multicolumn{2}{c}{HiP3 (Ours)} 
  & \multicolumn{2}{c}{AWAC} & 
  \multicolumn{2}{c}{PEX} & \multicolumn{2}{c}{Cal-QL} & 
  \multicolumn{2}{c}{MBPO} & \multicolumn{2}{c}{HiP3 (Ours)} \\
\midrule

\multirow{3}{*}{\rotatebox[origin=c]{90}{Random}}
 & HalfCheetah
    & \colorbox{white}{35.6}&{\color[HTML]{525252}$\pm$0.5}
    & \colorbox{white}{56.9}&{\color[HTML]{525252}$\pm$1.2}
    & \colorbox{white}{15.4}&{\color[HTML]{525252}$\pm$5.6}
    & \colorbox{white}{66.0}&{\color[HTML]{525252}$\pm$15.4}
    & \colorbox{mine}{82.6}&{\color[HTML]{525252}$\pm$2.3}
    & \colorbox{white}{33.2}&{\color[HTML]{525252}$\pm$0.4}
    & \colorbox{white}{42.2}&{\color[HTML]{525252}$\pm$2.6}
    & \colorbox{white}{11.1}&{\color[HTML]{525252}$\pm$2.5}
    & \colorbox{white}{47.1}&{\color[HTML]{525252}$\pm$22.1}
    & \colorbox{mine}{63.5}&{\color[HTML]{525252}$\pm$2.2} \\

 & Hopper
    & \colorbox{white}{11.7}&{\color[HTML]{525252}$\pm$0.2}
    & \colorbox{white}{22.2}&{\color[HTML]{525252}$\pm$4.1}
    & \colorbox{white}{12.7}&{\color[HTML]{525252}$\pm$8.5}
    & \colorbox{white}{98.2}&{\color[HTML]{525252}$\pm$6.1}
    & \colorbox{mine}{105.7}&{\color[HTML]{525252}$\pm$4.1}
    & \colorbox{white}{9.2}&{\color[HTML]{525252}$\pm$0.1}
    & \colorbox{white}{23.5}&{\color[HTML]{525252}$\pm$17.2}
    & \colorbox{white}{11.7}&{\color[HTML]{525252}$\pm$4.7}
    & \colorbox{white}{86.4}&{\color[HTML]{525252}$\pm$4.6}
    & \colorbox{mine}{95.2}&{\color[HTML]{525252}$\pm$5.5} \\

 & Walker2d
    & \colorbox{white}{12.2}&{\color[HTML]{525252}$\pm$6.7}
    & \colorbox{white}{9.8}&{\color[HTML]{525252}$\pm$2.2}
    & \colorbox{white}{1.6}&{\color[HTML]{525252}$\pm$1.3}
    & \colorbox{mine}{51.4}&{\color[HTML]{525252}$\pm$35.9}
    & \colorbox{white}{35.7}&{\color[HTML]{525252}$\pm$26.0}
    & \colorbox{white}{7.0}&{\color[HTML]{525252}$\pm$1.3}
    & \colorbox{white}{8.5}&{\color[HTML]{525252}$\pm$0.7}
    & \colorbox{white}{-0.6}&{\color[HTML]{525252}$\pm$1.1}
    & \colorbox{white}{44.6}&{\color[HTML]{525252}$\pm$11.4}
    & \colorbox{mine}{49.2}&{\color[HTML]{525252}$\pm$26.7} \\

\midrule

\multirow{3}{*}{\rotatebox[origin=c]{90}{\shortstack{Medium}}}
 & HalfCheetah
    & \colorbox{white}{43.6}&{\color[HTML]{525252}$\pm$0.8}
    & \colorbox{white}{44.1}&{\color[HTML]{525252}$\pm$0.3}
    & \colorbox{white}{49.7}&{\color[HTML]{525252}$\pm$0.3}
    & \colorbox{white}{66.0}&{\color[HTML]{525252}$\pm$15.4}
    & \colorbox{mine}{90.1}&{\color[HTML]{525252}$\pm$4.1}
    & \colorbox{white}{33.9}&{\color[HTML]{525252}$\pm$0.4}
    & \colorbox{white}{30.4}&{\color[HTML]{525252}$\pm$0.9}
    & \colorbox{white}{35.4}&{\color[HTML]{525252}$\pm$0.1}
    & \colorbox{white}{47.1}&{\color[HTML]{525252}$\pm$22.1}
    & \colorbox{mine}{69.3}&{\color[HTML]{525252}$\pm$2.1} \\

 & Hopper
    & \colorbox{white}{101.5}&{\color[HTML]{525252}$\pm$0.4}
    & \colorbox{white}{99.4}&{\color[HTML]{525252}$\pm$1.2}
    & \colorbox{white}{83.3}&{\color[HTML]{525252}$\pm$12.3}
    & \colorbox{white}{98.2}&{\color[HTML]{525252}$\pm$6.1}
    & \colorbox{mine}{106.7}&{\color[HTML]{525252}$\pm$3.0}
    & \colorbox{white}{88.5}&{\color[HTML]{525252}$\pm$0.1}
    & \colorbox{white}{74.8}&{\color[HTML]{525252}$\pm$15.9}
    & \colorbox{white}{37.6}&{\color[HTML]{525252}$\pm$4.6}
    & \colorbox{white}{86.4}&{\color[HTML]{525252}$\pm$4.6}
    & \colorbox{mine}{95.4}&{\color[HTML]{525252}$\pm$3.8} \\

 & Walker2d
    & \colorbox{white}{79.0}&{\color[HTML]{525252}$\pm$1.7}
    & \colorbox{white}{72.0}&{\color[HTML]{525252}$\pm$1.2}
    & \colorbox{mine}{83.1}&{\color[HTML]{525252}$\pm$2.2}
    & \colorbox{white}{51.4}&{\color[HTML]{525252}$\pm$35.9}
    & \colorbox{white}{76.4}&{\color[HTML]{525252}$\pm$16.0}
    & \colorbox{mine}{70.4}&{\color[HTML]{525252}$\pm$0.3}
    & \colorbox{white}{61.5}&{\color[HTML]{525252}$\pm$1.0}
    & \colorbox{white}{59.6}&{\color[HTML]{525252}$\pm$0.4}
    & \colorbox{white}{44.6}&{\color[HTML]{525252}$\pm$11.4}
    & \colorbox{white}{59.2}&{\color[HTML]{525252}$\pm$6.4} \\

\midrule

\multirow{3}{*}{\rotatebox[origin=c]{90}{\shortstack{Medium\\-replay}}}
 & HalfCheetah
    & \colorbox{white}{45.3}&{\color[HTML]{525252}$\pm$0.1}
    & \colorbox{white}{45.9}&{\color[HTML]{525252}$\pm$0.1}
    & \colorbox{white}{47.8}&{\color[HTML]{525252}$\pm$0.1}
    & \colorbox{white}{66.0}&{\color[HTML]{525252}$\pm$15.4}
    & \colorbox{mine}{89.6}&{\color[HTML]{525252}$\pm$2.7}
    & \colorbox{white}{29.9}&{\color[HTML]{525252}$\pm$3.8}
    & \colorbox{white}{30.4}&{\color[HTML]{525252}$\pm$3.5}
    & \colorbox{white}{32.6}&{\color[HTML]{525252}$\pm$0.2}
    & \colorbox{white}{47.1}&{\color[HTML]{525252}$\pm$22.1}
    & \colorbox{mine}{69.4}&{\color[HTML]{525252}$\pm$3.6} \\

 & Hopper
    & \colorbox{white}{29.6}&{\color[HTML]{525252}$\pm$2.8}
    & \colorbox{white}{36.0}&{\color[HTML]{525252}$\pm$7.0}
    & \colorbox{white}{95.1}&{\color[HTML]{525252}$\pm$1.4}
    & \colorbox{white}{98.2}&{\color[HTML]{525252}$\pm$6.1}
    & \colorbox{mine}{103.4}&{\color[HTML]{525252}$\pm$3.7}
    & \colorbox{white}{29.6}&{\color[HTML]{525252}$\pm$2.8}
    & \colorbox{white}{39.4}&{\color[HTML]{525252}$\pm$14.7}
    & \colorbox{white}{65.9}&{\color[HTML]{525252}$\pm$14.0}
    & \colorbox{white}{86.4}&{\color[HTML]{525252}$\pm$4.6}
    & \colorbox{mine}{100.1}&{\color[HTML]{525252}$\pm$11.0} \\

 & Walker2d
    & \colorbox{white}{34.5}&{\color[HTML]{525252}$\pm$8.4}
    & \colorbox{white}{40.6}&{\color[HTML]{525252}$\pm$7.3}
    & \colorbox{white}{85.5}&{\color[HTML]{525252}$\pm$4.9}
    & \colorbox{white}{51.4}&{\color[HTML]{525252}$\pm$35.9}
    & \colorbox{mine}{94.1}&{\color[HTML]{525252}$\pm$12.1}
    & \colorbox{white}{20.0}&{\color[HTML]{525252}$\pm$3.2}
    & \colorbox{white}{38.0}&{\color[HTML]{525252}$\pm$4.3}
    & \colorbox{white}{46.4}&{\color[HTML]{525252}$\pm$3.7}
    & \colorbox{white}{44.6}&{\color[HTML]{525252}$\pm$11.4}
    & \colorbox{mine}{77.3}&{\color[HTML]{525252}$\pm$10.4} \\

\midrule

\multirow{3}{*}{\rotatebox[origin=c]{90}{\shortstack{Medium\\-expert}}}
 & HalfCheetah
    & \colorbox{mine}{103.9}&{\color[HTML]{525252}$\pm$1.5}
    & \colorbox{white}{48.3}&{\color[HTML]{525252}$\pm$5.6}
    & \colorbox{white}{92.5}&{\color[HTML]{525252}$\pm$1.6}
    & \colorbox{white}{66.0}&{\color[HTML]{525252}$\pm$15.4}
    & \colorbox{white}{92.8}&{\color[HTML]{525252}$\pm$3.4}
    & \colorbox{mine}{77.6}&{\color[HTML]{525252}$\pm$0.7}
    & \colorbox{white}{29.9}&{\color[HTML]{525252}$\pm$1.0}
    & \colorbox{white}{64.4}&{\color[HTML]{525252}$\pm$1.9}
    & \colorbox{white}{47.1}&{\color[HTML]{525252}$\pm$22.1}
    & \colorbox{white}{74.7}&{\color[HTML]{525252}$\pm$2.0} \\

 & Hopper
    & \colorbox{mine}{112.3}&{\color[HTML]{525252}$\pm$0.4}
    & \colorbox{white}{89.0}&{\color[HTML]{525252}$\pm$31.6}
    & \colorbox{white}{112.2}&{\color[HTML]{525252}$\pm$0.2}
    & \colorbox{white}{98.2}&{\color[HTML]{525252}$\pm$6.1}
    & \colorbox{white}{106.5}&{\color[HTML]{525252}$\pm$4.7}
    & \colorbox{white}{88.8}&{\color[HTML]{525252}$\pm$4.4}
    & \colorbox{white}{50.9}&{\color[HTML]{525252}$\pm$9.5}
    & \colorbox{white}{88.0}&{\color[HTML]{525252}$\pm$5.5}
    & \colorbox{white}{86.4}&{\color[HTML]{525252}$\pm$4.6}
    & \colorbox{mine}{99.1}&{\color[HTML]{525252}$\pm$7.6} \\

 & Walker2d
    & \colorbox{white}{105.8}&{\color[HTML]{525252}$\pm$1.8}
    & \colorbox{white}{89.0}&{\color[HTML]{525252}$\pm$18.0}
    & \colorbox{mine}{109.6}&{\color[HTML]{525252}$\pm$0.7}
    & \colorbox{white}{51.4}&{\color[HTML]{525252}$\pm$35.9}
    & \colorbox{white}{79.3}&{\color[HTML]{525252}$\pm$10.6}
    & \colorbox{white}{50.6}&{\color[HTML]{525252}$\pm$8.5}
    & \colorbox{white}{54.3}&{\color[HTML]{525252}$\pm$2.8}
    & \colorbox{mine}{79.2}&{\color[HTML]{525252}$\pm$3.4}
    & \colorbox{white}{44.6}&{\color[HTML]{525252}$\pm$11.4}
    & \colorbox{white}{74.6}&{\color[HTML]{525252}$\pm$3.9} \\

\midrule
\multicolumn{2}{l}{Total Mean}
 & \colorbox{white}{59.6}&{\color[HTML]{525252}$\pm$2.2}
 & \colorbox{white}{54.4}&{\color[HTML]{525252}$\pm$6.6}
 & \colorbox{white}{65.7}&{\color[HTML]{525252}$\pm$3.3}
 & \colorbox{white}{71.9}&{\color[HTML]{525252}$\pm$19.1}
 & \colorbox{mine}{88.6}&{\color[HTML]{525252}$\pm$7.7}
 & \colorbox{white}{44.9}&{\color[HTML]{525252}$\pm$4.9}
 & \colorbox{white}{40.3}&{\color[HTML]{525252}$\pm$6.2}
 & \colorbox{white}{44.3}&{\color[HTML]{525252}$\pm$3.6}
 & \colorbox{white}{59.4}&{\color[HTML]{525252}$\pm$12.7}
 & \colorbox{mine}{77.3}&{\color[HTML]{525252}$\pm$7.1} \\

\bottomrule
\end{tabular}
}
\caption{Performance comparison on the same and novel online tasks. Novel online tasks are created by modifying the environments' gravity, friction, noise, and reward terms. Scores for each dataset are normalized according to the D4RL benchmark~\citep{fu2020d4rl}: $\textbf{Normalized Score} = 100\times\frac{\text{return}-\text{random return}}{\text{expert return}-\text{random return}}$. Full results can be found in Table~\ref{table:baseline_comparison_mujoco_appendix} in the Appendix.} 
\label{table:baseline_comparison}
\vspace{-3mm}
\end{table*}
\begin{figure*}[htbp]
    \centering
    \begin{subfigure}[b]{0.49\textwidth}
        \centering
        \includegraphics[width=\textwidth]{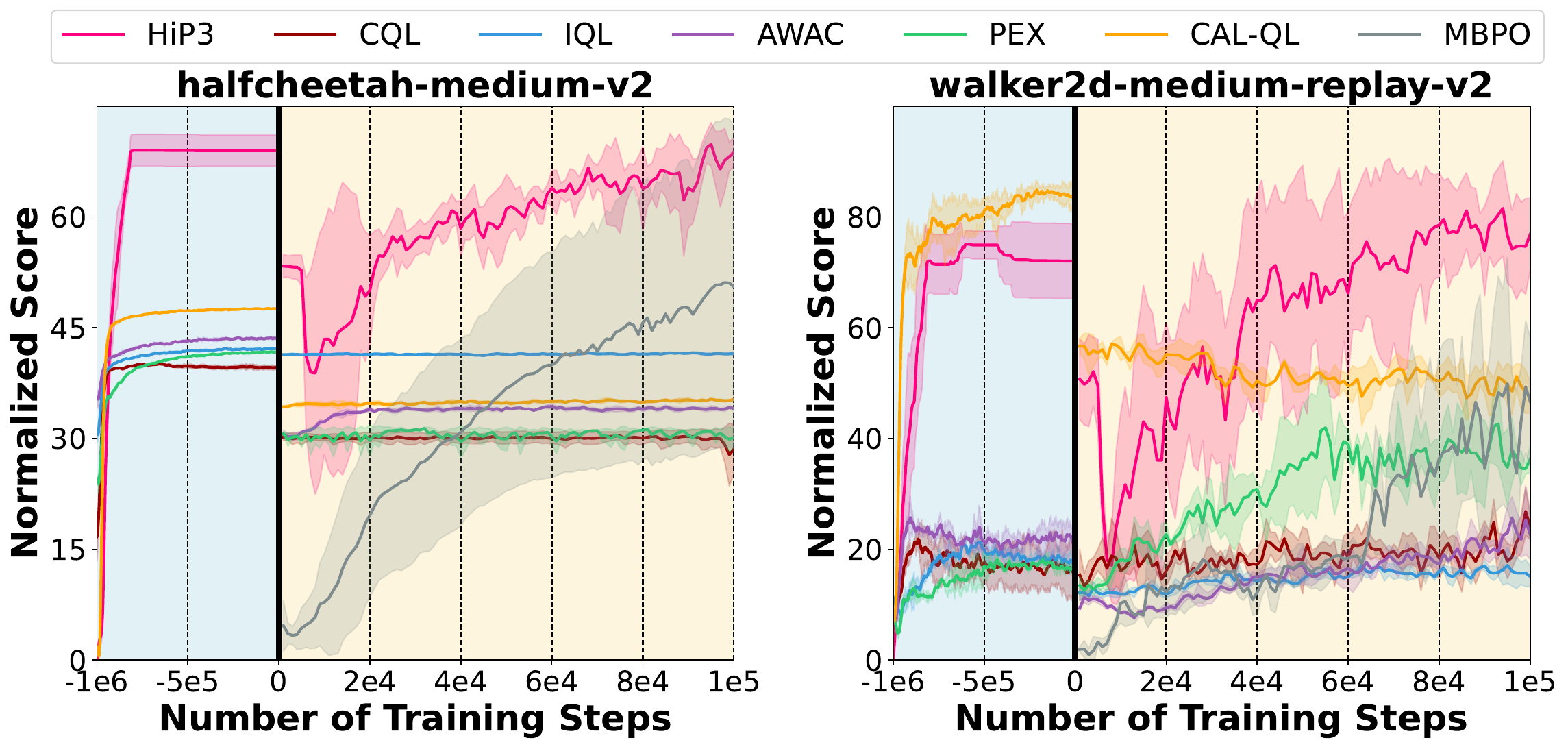}\vspace{-5pt}
        \caption{Learning curves on novel online tasks}
        \label{fig:baseline_comparison_example_sub2}
    \end{subfigure}
    \begin{subfigure}[b]{0.49\textwidth}
        \centering
        \includegraphics[width=\textwidth]{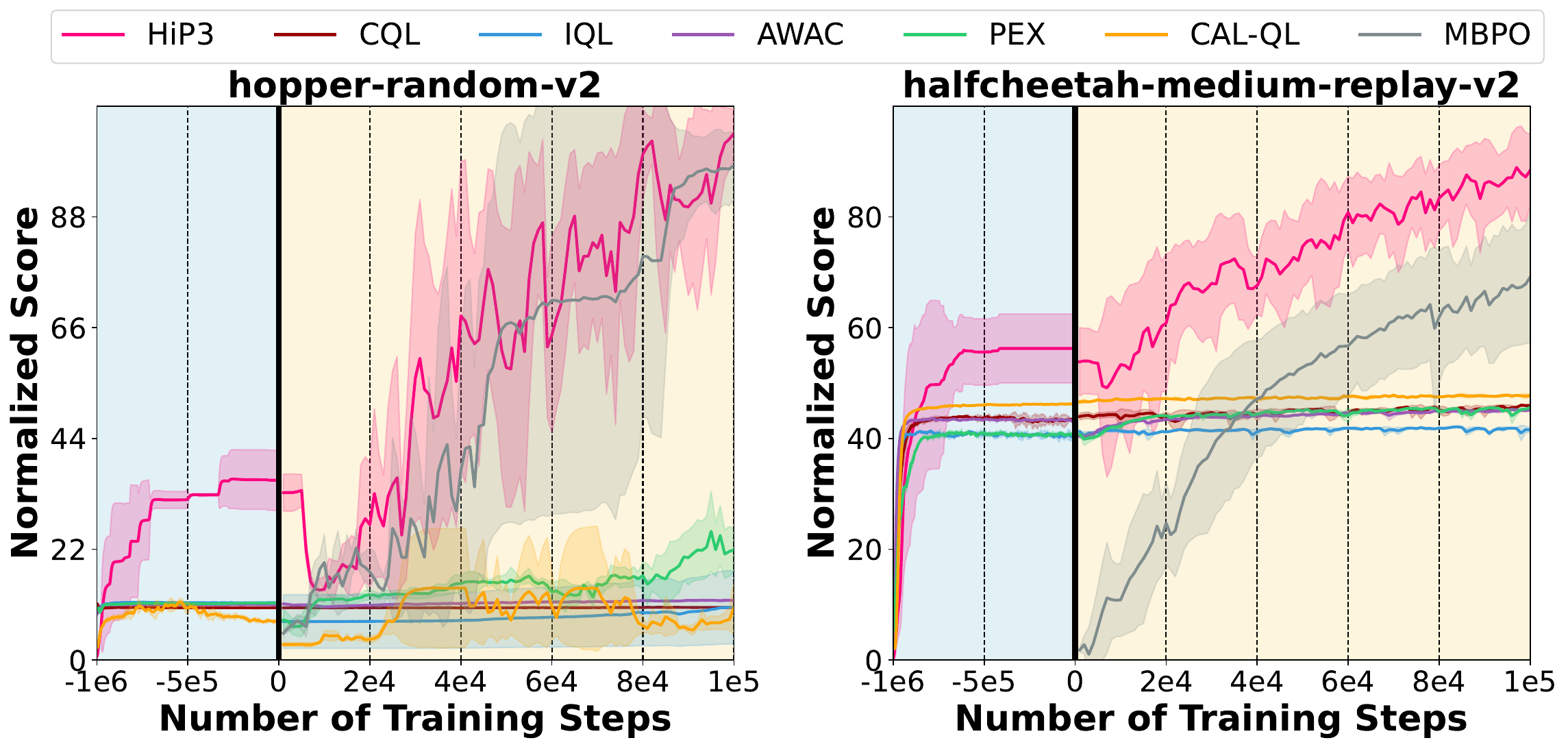}\vspace{-5pt}
        \caption{Learning curves on the same online tasks}
    \label{fig:baseline_comparison_example_sub1}
    \end{subfigure}
    \hfill\vspace{-5pt}
    \caption{Offline-to-online learning curves on the same and novel online tasks. The blue and yellow shaded regions denote offline pre-training and online fine-tuning, respectively. Curves are averaged over 5 random seeds with 10 evaluations per seed, and shaded areas show standard deviation. Full results across all datasets are available in Figure~\ref{fig:comparision_same_env} and Figure~\ref{fig:comparision_diff_env} in the Appendix.}
    \label{fig:baseline_comparison_example}
    \vspace{-2mm}
\end{figure*}

As shown in Table~\ref{table:baseline_comparison}, HiP3 achieves the strongest average performance across both the same-task and novel-task settings, with the best final performance on 7 of 12 same-task datasets and 9 of 12 novel-task datasets. Consistent with these aggregate results, Figure~\ref{fig:baseline_comparison_example} shows that several representative O2O baselines struggle on novel tasks and, in some cases, also underperform on same-task settings initialized from lower-quality offline data such as random or medium-replay datasets. For readability, the main table reports representative baselines, while complete per-task comparisons, including IQL, CQL, and the available FamO2O results, are given in Appendix Table~\ref{table:baseline_comparison_mujoco_appendix}.


Together, these results indicate that HiP3 improves adaptation across varied online tasks by combining a diverse offline policy pool with state-dependent online selection. Because different states favor different optimism-pessimism trade-offs, HiP3 can switch among candidate policies during fine-tuning, whereas fixed-trade-off baselines remain tied to the bias of a single offline policy.

To test whether HiP3 generalizes beyond MuJoCo locomotion, we also evaluate it on Adroit and Maze2D. HiP3 remains competitive on the Adroit tasks and achieves the best performance on Maze2D among the compared baselines, suggesting that the benefits of HiP3 may extend to additional continuous-control and navigation settings. Detailed results are provided in Appendix~\ref{app:add_domain}.



\vspace{-2mm}
\subsection{Adaptation under Stronger Distribution Shifts}
\vspace{-0.5mm}
We next evaluate HiP3 on a harder benchmark designed to stress adaptation under larger mismatches between the offline and online stages. Starting from the novel online tasks, we further modify both the environment dynamics and the reward design: we increase gravity from 11.5 to 13.5, scale the friction coefficients from $1.2\times$ to $1.25\times$, decrease the forward-reward weight from 0.85 to 0.8, and increase the control-cost weight from 0.2 to 0.25.

These modifications create substantially harder adaptation settings than the standard novel online tasks. We compare HiP3 with several recent O2O RL baselines, including SO2~\citep{zhang2024perspective}, WSRL~\citep{zhouefficient2025}, FQL~\citep{fql_park2025} and EDIS instantiated with IQL and Cal-QL backbones~\citep{liuenergy2024}, under the same 100K-step online fine-tuning budget using their official implementations or default settings. As shown in Table~\ref{tbl:more_challenging_setting}, HiP3 achieves the strongest average performance and remains competitive or superior on most tasks under these harder shifts. These results indicate that the combination of a diverse policy pool and state-dependent policy selection remains effective when the online task departs more substantially from the offline data distribution.

\begin{table}[htb]
\begin{center}
\small
\begin{tabular}{lcccccc}
\toprule
\textbf{Method} & \textbf{HiP3} & \textbf{SO2} & \textbf{WSRL} & \textbf{FQL} & \textbf{IQL-EDIS} & \textbf{Cal-QL-EDIS}\\
\midrule
hc-r-v2       & 42.9±3.9 & 29.8±0.7 & 38.7±1.5 & 22.2±1.8 & 15.7±0.6 & \textbf{48.9±4.6}\\
hc-mr-v2  & \textbf{55.8±4.6} & 35.9±0.5 & 52.4±0.5 & 44.1±0.3 & 27.2±0.3 & 43.0±2.1\\
hop-r-v2       & \textbf{65.4±6.3}  & 14.0±11.3 & 9.5±0.6 & 12.9±0.4 & 6.0±0.1 & 8.7±1.5\\
hop-mr-v2  & \textbf{102.1±4.1}  & 17.1±1.1 & 87.3±0.3 & 50.8±14.8 & 12.8±0.7 & 75.2±0.9\\
wk-r-v2       & \textbf{55.2±7.2}  & 39.5±2.8 & 53.6±7.6 & 5.7±1.8 & 3.5±0.3 & 27.0±11.9\\
wk-mr-v2  & 74.9±5.2  & 45.3±10.6 & \textbf{75.9±5.9} & 29.3±12.0 & 54.5±0.8 & 67.6±10.1\\
\midrule
mean & \textbf{66.1±5.2} & 30.3±4.5 & 52.9±2.7 & 27.5±5.2 & 20.0±0.5 & 45.1±5.2\\
\bottomrule
\end{tabular}\vspace{3pt}
\caption{Performance under more challenging distribution shifts. We modify environment dynamics and reward functions to create harder adaptation scenarios. (Notations: hc = halfcheetah, hop = hopper, wk = walker2d; r = random, mr = medium-replay.)}
\label{tbl:more_challenging_setting}
\end{center}
\vspace{-6mm}
\end{table}

\vspace{-2mm}
\subsection{Ablation Studies}
\begin{figure*}[t]
    \centering
    \begin{subfigure}[b]{0.497\textwidth}
        \centering
        \includegraphics[width=\textwidth]{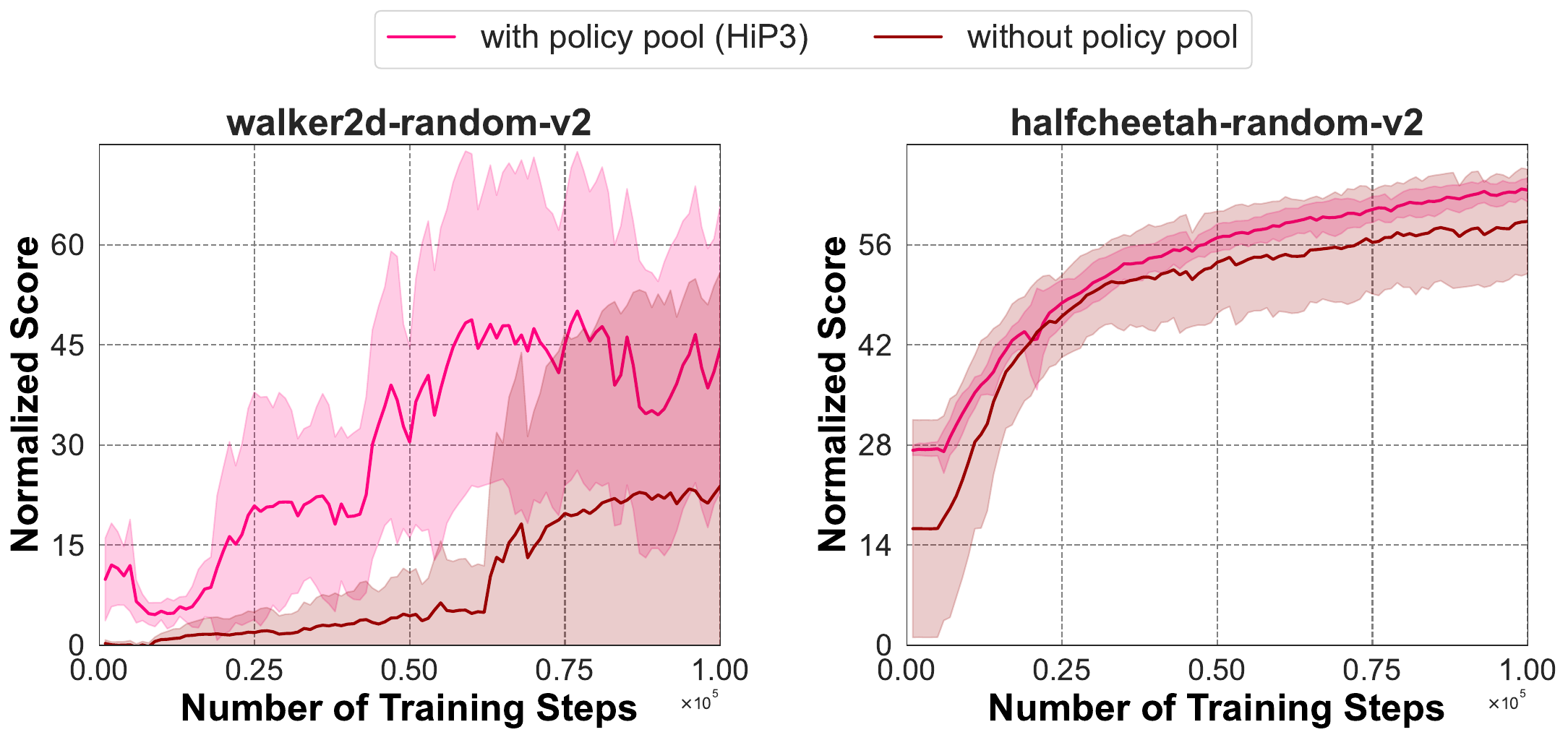}\vspace{-5pt}
        \caption{}
        \label{fig:ablation_policy_pool}
    \end{subfigure}
    \begin{subfigure}[b]{0.497\textwidth}
        \centering
        \includegraphics[width=\textwidth]{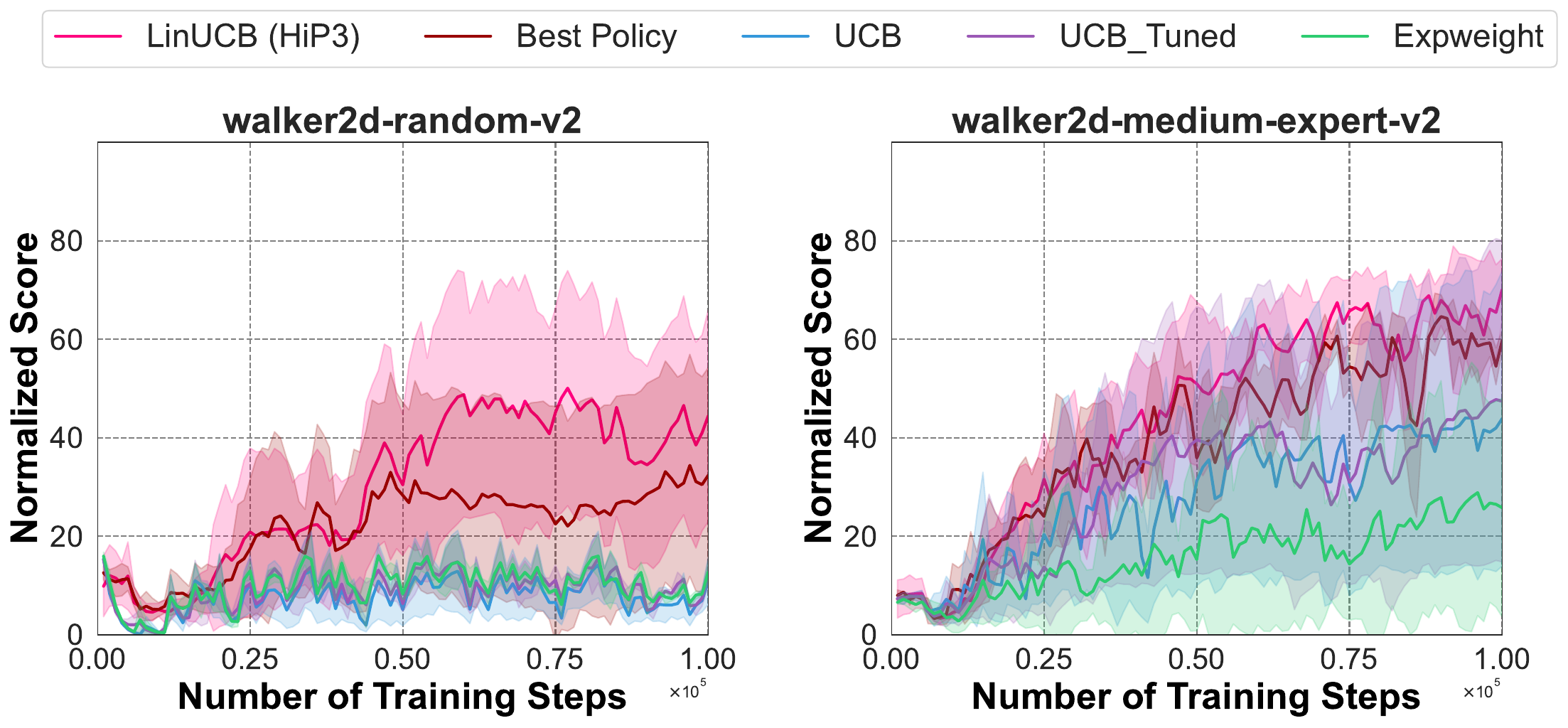}\vspace{-5pt}
        \caption{}
        \label{fig:ablation_select_algorithm}
    \end{subfigure}
    \caption{Ablations on novel online tasks: (a) removing the policy pool; (b) policy-selection algorithms. Full curves are provided in Figure~\ref{fig:ablation_policypool_appendix} and Figure~\ref{fig:ablation_select_algorithm_appendix} in the Appendix, respectively.}
    \label{fig:ablation_1}
    \vspace{-1mm}
\end{figure*}

\vspace{-0.5mm}
The preceding results establish HiP3’s adaptation advantage under both standard and stronger task shifts. We next isolate four design choices behind this performance: preserving multiple offline policies, state-dependent policy selection, MOSAIC-based pool construction, and candidate-set initialization. Figure~\ref{fig:ablation_1} examines the first two components, while Table~\ref{tbl:ablation} examines the latter two. Additional hyperparameter sensitivity analyses are reported in Appendix~\ref{app:hyper_sensitive}, while model and training sensitivity analyses are provided in Appendix~\ref{app:addition_ablation}.

Figure~\ref{fig:ablation_policy_pool} compares HiP3 with a single-policy variant that keeps the same Dyna-style online fine-tuning procedure but removes the offline policy pool. The multi-policy HiP3 framework achieves a higher mean score in all 12 reported novel online task settings. The complete 12-task comparison in Appendix Table~\ref{tab:single_policy_ablation} shows that preserving multiple offline trade-offs provides a stronger starting set for online adaptation.


Figure~\ref{fig:ablation_select_algorithm} shows the performance of HiP3 with various policy selection algorithms, including Best Policy, UCB~\citep{auer2002finite}, UCB-Tuned~\citep{auer2002finite}, Expweight~\citep{kessler2022same} and LinUCB~\citep{chu2011contextual}, which is the policy-selection algorithm used in HiP3. Holding the candidate pool fixed, LinUCB achieves the strongest performance on most evaluated tasks relative to fixed-policy and non-contextual selection alternatives. This pattern suggests that a single fixed policy is often insufficient during deployment because different states may favor different optimism-pessimism trade-offs. For instance, a pessimistic policy may be preferable in in-distribution states, whereas an optimistic policy may be more effective in under-explored states. These results support the effectiveness of state-dependent policy selection with LinUCB during online fine-tuning.

\begin{table*}[t]
\centering
\footnotesize
\setlength{\tabcolsep}{3pt}

\begin{subtable}[t]{0.35\textwidth}
\centering
\begin{tabular}{lcc}
\toprule
\textbf{Method} & \textbf{HiP3} & \textbf{HiP3-LS} \\
\midrule
halfcheetah-r-v2 
& 63.5$\pm$2.2 
& 47.5$\pm$17.1 \\

halfcheetah-mr-v2 
& 69.4$\pm$3.6 
& 57.8$\pm$15.4 \\

hopper-r-v2 
& 95.2$\pm$5.5 
& 62.3$\pm$21.9 \\

hopper-mr-v2 
& 100.1$\pm$11.0 
& 82.9$\pm$6.0 \\

\midrule
mean 
& 82.1$\pm$5.5 
& 62.6$\pm$15.1 \\

\bottomrule
\end{tabular}
\caption{}
\label{tbl:ablation_ls}
\end{subtable}
\hspace{0.04\textwidth}
\begin{subtable}[t]{0.58\textwidth}
\centering
\begin{tabular}{lcccc}
\toprule
\textbf{Method} 
& \textbf{HiP3 (Rollout)} 
& \textbf{HiP3 (FQE)} 
& \textbf{PEX} 
& \textbf{Cal-QL} \\
\midrule

hc-r-v2 
& 63.5$\pm$2.2 
& 58.7$\pm$1.7 
& 42.2$\pm$2.6 
& 11.1$\pm$2.5 \\

hc-mr-v2 
& 69.4$\pm$3.6 
& 67.8$\pm$3.3 
& 30.4$\pm$3.5 
& 32.6$\pm$0.2 \\

hop-r-v2 
& 95.2$\pm$5.5 
& 87.1$\pm$9.5 
& 23.5$\pm$17.2 
& 11.7$\pm$4.7 \\

hop-mr-v2 
& 100.1$\pm$11.0 
& 100.1$\pm$6.8 
& 39.4$\pm$14.7 
& 65.9$\pm$14.0 \\

\midrule
mean 
& 82.1$\pm$5.5 
& 78.4$\pm$5.3 
& 33.9$\pm$9.5 
& 30.3$\pm$5.4 \\

\bottomrule
\end{tabular}
\caption{}
\label{tbl:ablation_choosing_method}
\end{subtable}

\vspace{4pt}

\caption{
Ablation and controlled comparison studies.
(a) Comparison with a matched linear-scalarization baseline.
(b) Comparison of candidate-set initialization strategies, including target-interaction-free candidate ranking via FQE.
}
\label{tbl:ablation}
\end{table*}


The multi-policy ablation establishes the value of preserving multiple candidates. Table~\ref{tbl:ablation_ls} compares MOSAIC-based and fixed-scalarization pool-construction pipelines under matched policy-optimization budgets. HiP3-LS uses the same model-based backbone, offline data, total policy-optimization budget, and online selection framework, while training five policies with $wJ^{\hat r}(\pi_\theta)+(1-w)J^{\hat u}(\pi_\theta)$, where $w\in\{0.9,0.7,0.5,0.3,0.1\}$. Under this controlled protocol, HiP3 consistently outperforms HiP3-LS, supporting the advantage of MOSAIC-based policy-pool construction.

Table~\ref{tbl:ablation_choosing_method} examines whether HiP3 remains effective when candidate initialization avoids target-task interaction. Replacing rollout-based ranking with Fitted Q Evaluation (FQE)~\citep{le2019batch} avoids target-environment interactions during candidate ranking, with only a modest reduction in subsequent adaptation performance. HiP3+FQE still substantially outperforms PEX and Cal-QL in these settings, demonstrating that HiP3 retains strong adaptation performance even with interaction-free candidate initialization.

\subsection{Online Policy-Selection Dynamics}
\vspace{-0.5mm}

\begin{figure*}[h]
    \centering
    \includegraphics[width=1.0\linewidth]{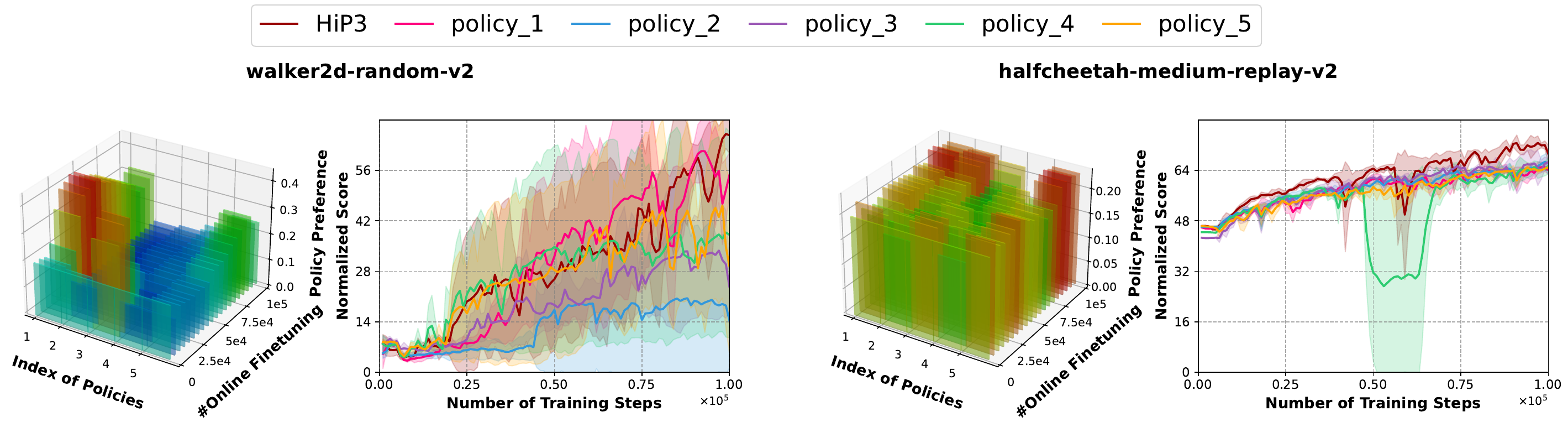}\vspace{-5pt}
    \caption{Heat maps visualize the policy preferences $P_t$, obtained by softmax-normalizing the LinUCB scores. HiP3 executes $k_t=\arg\max_k P_t(k)$ at each interaction step. A sharply peaked preference profile indicates a clear preference for one candidate, whereas a diffuse profile indicates smaller preference margins among candidates. The accompanying curves compare HiP3 with the five candidate policies on two representative novel online tasks. Additional quantitative and qualitative analyses are provided in Appendix~\ref{app:linucb_analysis}.}
    \label{fig:effectiveness}
    \vspace{-1mm}
\end{figure*}

The preceding ablations establish the value of both policy diversity and state-dependent policy selection. We therefore inspect how LinUCB's preferences over the candidate pool evolve during online adaptation. Figure~\ref{fig:effectiveness} illustrates two representative patterns. On walker2d-random-v2, LinUCB develops a stronger preference for one candidate, and HiP3 closely tracks or slightly exceeds that candidate's performance. On halfcheetah-medium-replay-v2, several candidates retain comparable preferences, while HiP3 outperforms the fixed-candidate baselines. These examples illustrate how HiP3 uses the policy pool during adaptation and complement the policy-selection ablation in Figure~\ref{fig:ablation_select_algorithm}.

Table~\ref{tab:linucb_frequency} complements these preference visualizations by summarizing the policies actually executed across all 12 novel online task settings. The dominant-policy selection frequency varies across domains and tasks, and no single candidate accounts for the entire online run. Thus, a pronounced preference for one candidate should not be interpreted as exclusive execution of that policy. Additional policy-preference visualizations and a comparison of bandit model capacity are provided in Appendix~\ref{app:linucb_analysis}.

\begin{table}[H]
\centering
\footnotesize
\setlength{\tabcolsep}{6pt}
\begin{tabular}{lcc}
\toprule
\textbf{Domain} & 
\textbf{Mean Dominant-policy Selection Frequency} &
\textbf{Range Across Settings}\\
\midrule
HalfCheetah & 37.6\% & 28.6\%-62.0\%\\
Hopper & 35.1\% & 29.0\%-44.4\%\\
Walker2d & 54.4\% & 46.1\%-58.8\%\\
\midrule
Overall & 42.3\% & 28.6\%-62.0\%\\
\bottomrule
\end{tabular}\vspace{5pt}
\caption{
Quantitative analysis of LinUCB policy selection behavior on 12 novel online task settings.
The dominant policy is defined as the candidate selected most frequently over the full online run. Its selection frequency is the fraction of online steps assigned to that candidate.
}
\label{tab:linucb_frequency}
\end{table}

\vspace{-5pt}
\section{Conclusion}\label{sec: conclusion}
\vspace{-2pt}

This paper studies the optimism-pessimism trade-off in model-based O2O RL and introduces HiP3, which preserves diverse trade-offs offline and performs state-dependent policy selection online rather than committing to a single balance in advance. HiP3 formulates optimism and pessimism as two explicit objectives, uses MOSAIC to construct a Pareto policy pool, and uses a contextual bandit to select candidate policies conditioned on the current state throughout model-based fine-tuning. Across both same and novel online tasks, HiP3 achieves strong aggregate performance and maintains its advantage under stronger distribution shifts. These results support a simple design principle for model-based O2O RL: preserve trade-off diversity offline and defer the choice among candidate policies to online adaptation.


\textbf{Limitation:} While HiP3 performs strongly on the evaluated O2O RL benchmarks, we identify three key areas for further investigation. First, similar to prior methods, HiP3 still exhibits an initial performance degradation at the early stage of online fine-tuning. We provide an empirical analysis of this behavior in Appendix ~\ref{appendix:performance_drop} and believe that mitigating such transient instability remains a promising avenue for future work. Second, while our evaluation primarily focuses on the widely-adopted D4RL MuJoCo suite to ensure reproducibility and fair comparison, extending the proposed framework to higher-dimensional environments with more pronounced distribution shifts represents a valuable future direction. Third, because reward and uncertainty are model-derived surrogates, offline Pareto trade-offs may not exactly match real-environment trade-offs under severe model bias.



\looseness-1

\begin{ack}
\textbf{Funding.}
This work is partially supported by the CityUHK Startup Grant (Grant No. 9382014) and the Jiangxi Provincial Natural Science Foundation (Grant No. 20262BAC200479).

\textbf{Competing Interests.}
The authors declare no competing interests.
\end{ack}


{
\small
\bibliographystyle{unsrtnat}
\bibliography{ref}
}

\newpage
\appendix
\onecolumn
\section*{Organization of Appendix}
The appendix is organized as follows. In Section~\ref{app:limitation&impact}, we discuss the potential societal impacts of our work. In Section~\ref{app:theory_and_algo}, we provide additional theoretical and algorithmic details of HiP3, including the convergence analysis of MOSAIC, uncertainty estimation, the environment model, background on multi-objective optimization, comparison with linear scalarization, and the bandit selection mechanism. In Section~\ref{app:exp_setup}, we describe the experimental setup, implementation details, computing infrastructure and cost, and hyper-parameters used in our experiments. In Section~\ref{app:add_emp_result}, we present additional empirical analyses, including validation of the optimism-pessimism crossover behavior, results on additional domains, transient performance analysis, hyperparameter sensitivity studies, additional ablations, policy-selection analyses, complete benchmark results, and extended baseline comparisons.

\section{Societal Impacts}\label{app:limitation&impact}

This work contributes to offline-to-online reinforcement learning by explicitly modeling the trade-off between optimistic exploration and pessimistic conservatism, a central challenge when adapting policies under distribution shift. By enabling flexible policy selection during online interaction, the proposed approach has the potential to reduce unnecessary exploration and improve sample efficiency, which may benefit real-world learning systems where online interaction is costly or risky.

At the same time, as with many offline-to-online RL methods, transient performance degradation during early online fine-tuning highlights the need for cautious deployment, particularly in safety-critical settings or under severe distribution shifts. While our experiments focus on standard continuous control benchmarks, applying the proposed framework in real-world systems would require careful validation, monitoring, and safeguards to ensure stable behavior. We hope this work encourages further research into principled uncertainty-aware adaptation mechanisms that balance performance and safety in online learning.

\section{Theory and Algorithmic Details}\label{app:theory_and_algo}


\subsection{Convergence Analysis}\label{appendix: convergence analysis}
\textbf{Assumption 1.} Suppose the $m$ objectives $f_1, f_2, \ldots, f_m$ are differentiable and their gradients are $L_i$-Lipschitz continuous with constant $L_i > 0$.

\textbf{Assumption 2.}  For any $i\in \{1,2,\cdots, m\}$, $\{g_i(\theta_t)\}_{t=1}^{\infty}$ is an independent stochastic gradient sequence satisfying:
$$\mathbb{E}[g_i(\theta_t)]=\nabla f_i(\theta_t),\ Var(g_i(\theta_t))=\mathbb{E}[||g_i(\theta_t)-\nabla f_i(\theta_t)||^2] \leq \sigma^2.$$

\textbf{MGDA Update:} In the implementation, MOSAIC updates the policy by minimizing a convex combination of three surrogate losses. For the analysis below, we work in the equivalent objective-space view and let $f_1,\ldots,f_m$ denote the corresponding objectives, so that descent on each surrogate loss is equivalent to ascent on its paired objective. In the ascent stage of MOSAIC (see Alg.~\ref{algo:mo_sac}), define the combined objective-space update direction
\[
d_t = \sum_{i=1}^m \beta_{i,t}\ g_i(\theta_t),
\]

where the weights $\beta_{i,t} \ge 0$ are chosen by solving the MGDA subproblem (Eq.~\ref{eq:4_5}), ensuring that no individual objective is decreased at that step (i.e., $\nabla f_i(\theta_t)^\top d_t \ge 0, \forall i$). The stochastic policy update in objective space is then:
\[\theta_{t+1} = \theta_t + \eta d_t.\]
Let $\bar{d}_t = \mathbb{E}_t[d_t]$ denote the conditional expected update direction.


\begin{theorem}[First-order convergence of the ascent stage]
Suppose that Assumptions 1 and 2 hold, and let $L=\max_i L_i$. Assume further that, for each objective $f_i$, there exists a finite upper reference value $f_i^{\star}$ such that $f_i(\theta_t) \leq f_i^{\star}$ for all iterates generated by the ascent stage. Define
\[
\Delta_i = f_i^{\star} - f_i(\theta_0), \qquad \Delta = \max_i\{\Delta_i\}.
\]
For any fixed $T$, choose the constant stepsize
\[
\eta=\frac{c}{\sqrt{T}}, \qquad 0<c<\frac{2}{L}.
\]
Then
\[
\frac{1}{T}\sum_{t=0}^{T-1}\mathbb{E}[\|\bar{d}_t\|^2]
= O\left(\frac{\Delta}{c\sqrt{T}}+\frac{cL\sigma^2}{\sqrt{T}}\right).
\]
Consequently, for any $\epsilon > 0$, after
\[
T=O\left(\frac{1}{\epsilon^2}\left(\frac{\Delta}{c}+cL\sigma^2\right)^2\right)
\]
iterations, we have
\[
\frac{1}{T}\sum_{t=0}^{T-1}\mathbb{E}[\|\bar{d}_t\|^2] \leq \epsilon.
\]
Thus, the ascent stage enjoys a first-order convergence guarantee toward Pareto stationarity in expectation.
\label{theorem1}
\end{theorem}

\begin{proof}
By $L_i$-smoothness and the ascent update $\theta_{t+1}=\theta_t+\eta d_t$, we have for each $f_i$:
\begin{align*}
f_i(\theta_{t+1}) &\geq f_i(\theta_t) + \nabla f_i(\theta_t)^\top(\theta_{t+1}-\theta_t) - \frac{L_i}{2} \|\theta_{t+1} - \theta_t\|^2\\
&=f_i(\theta_t) +\eta\ \nabla f_i(\theta_t)^\top d_t - \frac{L_i}{2}\eta^2 \|d_t\|^2.
\end{align*}
Taking conditional expectation $\mathbb{E}_t[\cdot]$ on both sides yields
\begin{align}\label{equ:ascent inequality}
\mathbb{E}_t[f_i(\theta_{t+1})]\geq f_i(\theta_t) +\eta\ \nabla f_i(\theta_t)^\top \bar{d}_t - \frac{L_i}{2}\eta^2 \mathbb{E}_t[\|d_t\|^2],
\end{align}
where $\bar{d}_t = \mathbb{E}_t[d_t]$.

Also,
\begin{equation}\label{equ:variance estimate}
\begin{split}
    \mathbb{E}_t\left[\|d_t\|^2\right]
    &=\left\|\bar{d}_t\right\|^2+\mathbb{E}_t\left[\left\|d_t-\bar{d}_t\right\|^2\right]\\
    &\leq\left\|\bar{d}_t\right\|^2+\sigma^2\sum_{i=1}^m\beta_{i,t}^2\\
    &\leq\left\|\bar{d}_t\right\|^2+\sigma^2\left(\sum_{i=1}^m\beta_{i,t}\right)^2\\
    &=\left\|\bar{d}_t\right\|^2+\sigma^2,
\end{split}
\end{equation}
where we use Assumption 2 and the independence of the stochastic gradients. The second inequality uses the fact
\[
\sum_{i=1}^m\beta_{i,t}^2\leq \left(\sum_{i=1}^m\beta_{i,t}\right)^2
\]
along with $\sum_{i=1}^m\beta_{i,t}=1$.

Substituting Eq.~(\ref{equ:variance estimate}) into Eq.~(\ref{equ:ascent inequality}) gives
\begin{align*}
 \mathbb{E}_t[f_i(\theta_{t+1})]&\geq f_i(\theta_t) +\eta\ \nabla f_i(\theta_t)^\top \bar{d}_t - \frac{L_i}{2}\eta^2 \left(\left\|\bar{d}_t\right\|^2+\sigma^2\right)\\
 &\geq f_i(\theta_t) +\eta\ \| \bar{d}_t\|^2 - \frac{L_i}{2}\eta^2 \left(\left\|\bar{d}_t\right\|^2+\sigma^2\right),
\end{align*}
where we use the fact $\nabla f_i^\top \bar d_t \ge \|\bar d_t\|^2$ (Lemma 2.1 in \citep{desideri2012multiple}) in the second inequality.

Rearranging the above inequality and then taking total expectation yields
\begin{align*}
\left(\eta-\frac{L_i}{2}\eta^2\right)\mathbb{E}[\| \bar{d}_t\|^2]\leq \mathbb{E}\left[f_i(\theta_{t+1})- f_i(\theta_{t})\right]+\frac{L_i}{2}\eta^2 \sigma^2.
\end{align*}
Summing over $t=0$ to $T-1$ and dividing both sides by $T$ gives
\begin{align*}
        \frac{1}{T}\sum_{t=0}^{T-1}\mathbb{E}\left[\| \bar{d}_t\|^2\right]\leq \frac{\mathbb{E}[f_i(\theta_T)]-f_i(\theta_0)}{T\left(\eta-\frac{L_i}{2}\eta^2\right)}+\frac{L_i\eta^2\sigma^2}{2\left(\eta-\frac{L_i}{2}\eta^2\right)}.
\end{align*}
Since $\mathbb{E}[f_i(\theta_T)] \leq f_i^{\star}$ by assumption, it follows that
\begin{align*}
    \frac{1}{T}\sum_{t=0}^{T-1}\mathbb{E}\left[\| \bar{d}_t\|^2\right]\leq \frac{\Delta_i}{T\left(\eta-\frac{L_i}{2}\eta^2\right)}+\frac{L_i\eta^2\sigma^2}{2\left(\eta-\frac{L_i}{2}\eta^2\right)}.
\end{align*}
Setting $\eta=\frac{c}{\sqrt{T}}$ with $0<c<\frac{2}{L}$ and using $L=\max_i L_i$, we obtain
\begin{align*}
    \frac{1}{T}\sum_{t=0}^{T-1}\mathbb{E}\left[\| \bar{d}_t\|^2\right]
    = O\left(\frac{\Delta}{c\sqrt{T}}+\frac{cL\sigma^2}{\sqrt{T}}\right).
\end{align*}
Therefore, for any $\epsilon>0$, after
\[
T=O\left(\frac{1}{\epsilon^2}\left(\frac{\Delta}{c}+cL\sigma^2\right)^2\right)
\]
iterations, we have
\[
\frac{1}{T}\sum_{t=0}^{T-1}\mathbb{E}\left[\| \bar{d}_t\|^2\right]\leq \epsilon.
\]
\end{proof}
\subsection{Uncertainty Estimation $u$ in HiP3}\label{appendix: uncertainty estimation}
The predicted uncertainty $u$ of the environment model $M$ is estimated as the maximum standard deviation among the ensemble, i.e., $u(s_t, a_t) = \max_{i=1}^N ||\sigma_i(s_t, a_t)||$. This ensemble-based uncertainty estimation strategy follows established practices in model-based offline RL such as MOPO~\citep{yu2020mopo} and MOReL~\citep{kidambi2020morel}. This method is favored for its scalability, ease of implementation, and effectiveness in representing epistemic uncertainty in deep neural networks. Although alternative techniques such as Bayesian methods~\citep{lin2023uncertainty} and unsupervised representation-based approaches~\citep{perez2023discrete} are available, recent studies~\citep{rahaman2021uncertainty, ghasemipour2022so} have demonstrated that deep ensembles tend to provide more accurate and better-calibrated uncertainty estimates. These results indicate that ensemble methods strike an effective balance between predictive performance and computational cost, making them a reliable and practical choice for uncertainty estimation in HiP3.

A limitation of using predicted return and uncertainty directly is that they may not match the true dynamics in out-of-distribution regions, especially when the offline dataset has limited coverage. Conservative approaches~\citep{yu2020mopo, rafailov2023moto} address this by penalizing high-uncertainty predictions. While effective in reducing overestimation, this requires careful tuning of penalty coefficients and may restrict exploration during online fine-tuning. 

HiP3 avoids the use of fixed penalties by formulating optimism and pessimism as two separate objectives in a bi-objective optimization problem. This formulation produces a Pareto set of offline policies with different optimism-pessimism levels. During online fine-tuning, HiP3 performs state-dependent policy selection over these policies.

\subsection{Environment Model $M$}\label{appendix: environmental model}
The environment model used by HiP3 consists of a bootstrap ensemble of $N$ environment models: $M=\{M_i\}_{i=1}^N$. Each member $i$ of the ensemble is a probabilistic neural network, taking state-action pairs $(s_t, a_t)$ as input and outputting a Gaussian distribution over the next state and reward: $\hat{P}_i(s_{t+1}, r|s_t, a_t) = \mathcal{N}(\mu_i(s_t, a_t), \sigma_i(s_t,a_t))$. The predicted reward is the ensemble mean of the reward components $\hat r(s_t,a_t)=\frac{1}{N}\sum_{i=1}^{N}\mu_{i,r}(s_t,a_t),$
where $\mu_{i,r}(s_t,a_t)$ denotes the reward component of the joint predicted mean $\mu_i(s_t,a_t)$. The predicted uncertainty $u$ of the environment model $M$ is estimated as the maximum standard deviation among the ensemble, i.e., $u(s_t, a_t) = \max_{i=1}^N ||\sigma_i(s_t, a_t)||$.

This environment model is trained according to the design of MOPO~\citep{yu2020mopo}. Specifically, we trained an ensemble of 7 models, from which the 5 best models ($N=5$) in terms of the prediction errors are selected. Each model of the ensemble is independently trained using the maximum likelihood estimation with mini-batch stochastic gradient descent~\citep{jain2018parallelizing}. During the inference stage, the policy interacts with a randomly selected model from the $N$ ensemble models, generating state-reward pairs from the corresponding distributions. This approach introduces variability by sampling different transitions from different models within a single episode, leading to a highly stochastic environment model. Previous works~\citep{janner2019trust, yu2020mopo} have shown that this environment model effectively mitigates the model exploitation problem in offline RL scenarios.

The hyper-parameters of the environment model $M$ are provided in Table~\ref{tbl:model_hyper-parameters}. For the MuJoCo tasks, we adopt the standard MBPO/MOPO world-model configuration~\citep{janner2019trust, yu2020mopo} given their proven effectiveness in similar continuous control environments.

\begin{table}[htb]
\begin{center}
\begin{tabular}{ll}
\toprule
\textbf{Hyper-parameter} & \textbf{Value}  \\
\midrule
Ensemble size & 7\\
Number of elite models & 5 \\
Model architecture & FC(200, 200, 200, 200) \\
Batch size & 256 \\
Optimizer & Adam\\
Learning rate & $1\times10^{-3}$\\
Holdout ratio & 0.2\\
\bottomrule
\end{tabular}\vspace{3pt}
\caption{The hyper-parameters used for environment model $M$}
\label{tbl:model_hyper-parameters}
\end{center}
\end{table}

\subsection{Multi-objective Optimization}~\label{app:moo}
Multi-objective optimization (MOO) considers the simultaneous optimization of
two or more potentially conflicting objectives~\citep{deb2002fast}. Formally,
given a decision space $\mathcal X$ and $d$ objective functions
$f_i:\mathcal X\rightarrow\mathbb R$ for $i\in\{1,\ldots,d\}$, we consider the
following vector optimization problem:
\[
\max_{x\in\mathcal X}\;
\mathbf f(x)
\triangleq
\bigl(f_1(x),f_2(x),\ldots,f_d(x)\bigr).
\]
The trade-offs among objectives arise because improving one objective may
degrade another, so a single solution that simultaneously maximizes all
objectives need not exist. MOO may therefore admit multiple Pareto-optimal
solutions representing different trade-offs. We use the following standard
definitions.

\paragraph{Dominance.}
Given two solutions $x_1,x_2\in\mathcal X$, we say that $x_1$ dominates
$x_2$ (denoted $x_1\succ x_2$) if
\[
f_i(x_1)\ge f_i(x_2), \quad \forall i\in\{1,\ldots,d\},
\]
and
\[
f_j(x_1)>f_j(x_2)
\]
for at least one objective $j$.

\paragraph{Pareto optimality.}
A solution $x^\star\in\mathcal X$ is Pareto-optimal if there does not exist
another solution $x'\in\mathcal X$ such that $x'\succ x^\star$.

\paragraph{Pareto set and Pareto front.}
The Pareto set is the set of Pareto-optimal solutions in the decision space:
\[
\mathcal P^\star
=
\{x\in\mathcal X
\mid
\nexists\,x'\in\mathcal X
\text{ such that }x'\succ x\}.
\]
The corresponding Pareto front is the image of the Pareto set in the objective
space:
\[
\mathcal F^\star
=
\{\mathbf f(x)\mid x\in\mathcal P^\star\}.
\]

By providing Pareto-optimal solutions spanning different regions of the
Pareto front, MOO enables different trade-offs to be preserved for subsequent
selection according to task requirements.

\subsection{Comparison with linear scalarization}~\label{app:ls_comparison}
Linear scalarization (LS) provides a natural baseline for multi-objective RL by combining multiple objectives through a fixed scalarization weight. In the context of model-based O2O RL, fixed linear scalarization introduces two structural limitations that are particularly relevant to optimism-pessimism trade-offs:


First, linear scalarization can recover supported Pareto-optimal solutions by varying the scalarization weight, but it cannot in general recover unsupported Pareto-optimal solutions when the attainable objective set is non-convex~\citep{boyd2004convex}. HiP3 instead uses reference-vector-guided constrained optimization together with MGDA, and is therefore not restricted to the supported solutions induced by fixed linear scalarization weights. This allows MOSAIC to target trade-off regions that linear scalarization may miss.

Second, a single LS objective fixes a global weight $\lambda$ and therefore a single optimism-pessimism balance. This can be extended by training multiple LS policies with different weights and selecting among them online, as in our HiP3-LS baseline. However, each candidate remains tied to a predefined scalarization weight, and different weights can map to the same or similar supported solutions, which can produce redundant candidates and limit the diversity available for online selection. HiP3 instead directly targets distinct trade-off regions through reference vectors and then performs state-dependent policy selection over the resulting pool.


For a controlled comparison, we construct HiP3-LS by replacing MOSAIC with fixed linear scalarization objectives. HiP3-LS trains five policies by maximizing
$$
wJ^{\hat r}(\pi_\theta)+(1-w)J^{\hat u}(\pi_\theta),
\qquad
w\in\{0.9,0.7,0.5,0.3,0.1\},
$$
which matches the first coordinates of HiP3's reference vectors. Each policy receives an equal share of the 1M-step offline budget and is used directly as one arm of the same LinUCB selector.

To provide an additional comparison with an existing multi-objective RL approach beyond linear scalarization, we also combine our bandit selector with DiME~\citep{abdolmaleki2021multi}. Since no open-source implementation is available, we implement DiME on top of AWAC following the original paper.

\begin{table}[htb]
\begin{center}
\small
\begin{tabular}{lcccc}
\toprule
\textbf{Method} & \textbf{HiP3} & \textbf{HiP3-LS} & \textbf{DiME+AWAC} & \textbf{AWAC} \\
\midrule
halfcheetah-random-v2       & 63.5±2.2 & 47.5±17.1 & 28.4±1.6 & 33.2±0.4\\
halfcheetah-medium-replay-v2  & 69.4±3.6 & 57.8±15.4 & 36.0±4.1 & 29.9±3.8\\
hopper-random-v2       & 95.2±5.5 & 62.3±21.9 & 7.0±4.9 & 9.2±0.1  \\
hopper-medium-replay-v2  & 100.1±11.0 & 82.9±6.0 & 51.2±8.0 & 29.6±2.8 \\
\midrule
mean & 82.1±5.5 & 62.6±15.1 & 30.7±4.7 & 25.5±1.8\\
\bottomrule
\end{tabular}\vspace{3pt}
\caption{Comparison with Linear Scalarization Method}
\label{tbl:ls_comparison}
\end{center}
\end{table}

Under the matched protocol, HiP3 achieves a mean score of 82.1 compared with 62.6 for HiP3-LS. DiME+AWAC improves over vanilla AWAC on several novel online tasks, while HiP3 achieves substantially higher aggregate performance under the same offline and online optimization budgets. These results support the motivation of Eq.~\ref{eq:3_2}: avoiding restriction to fixed linear scalarization enables more flexible offline trade-off coverage and provides a stronger policy pool for subsequent online adaptation.

\subsection{Contextual Policy Selection with LinUCB}~\label{app:bandit}
\begin{figure*}
    \centering
    \includegraphics[width=0.78\linewidth]{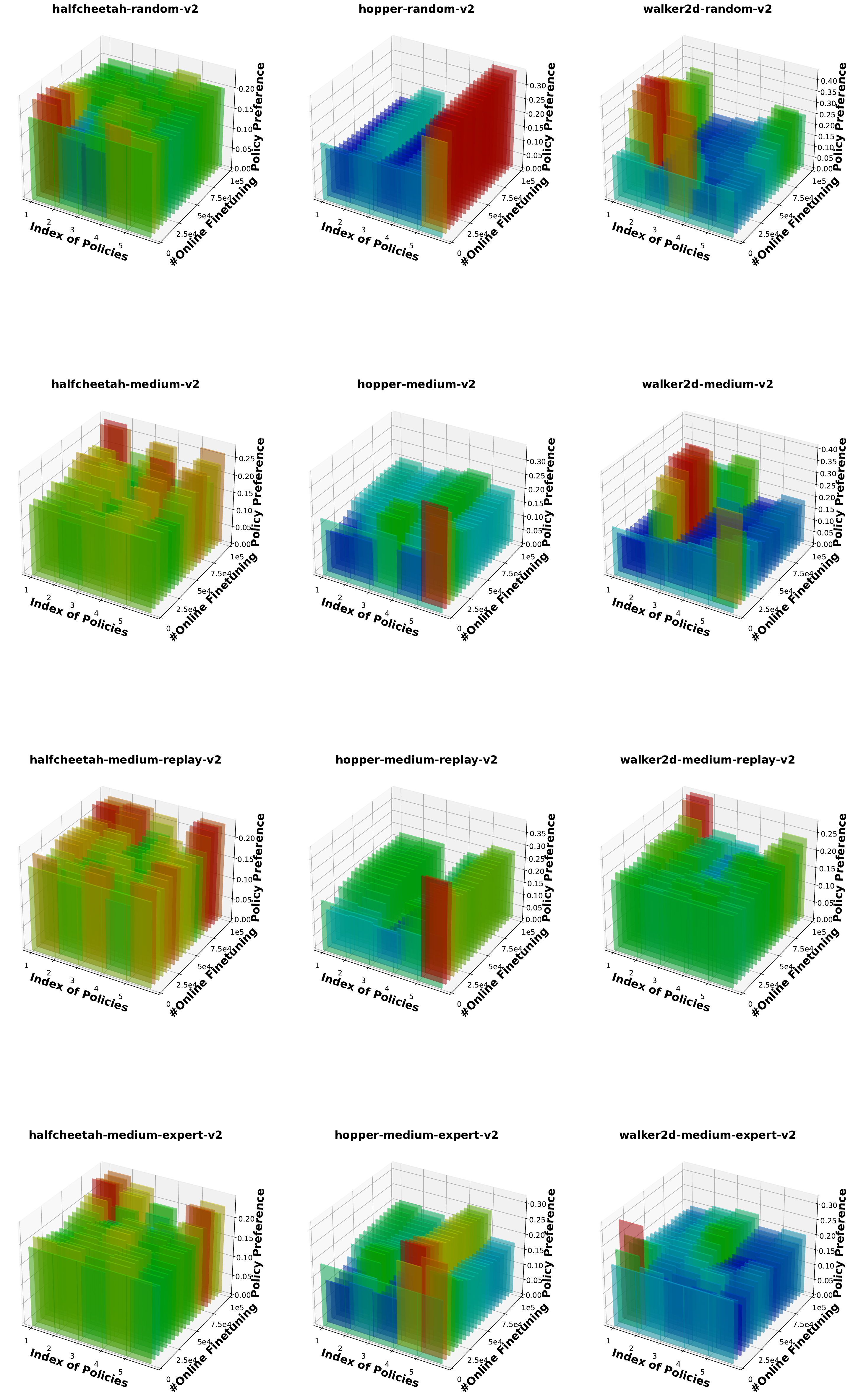}
    \caption{Online policy-selection dynamics across the novel online tasks. The heat maps visualize the policy preferences $P_t$ obtained by softmax-normalizing the LinUCB scores at each online interaction step. The visualizations show whether the selector develops a concentrated preference for one candidate or maintains comparable preferences across multiple candidates during adaptation.}
    \label{fig:effectiveness_appendix}
\end{figure*}

We use LinUCB~\citep{chu2011contextual} as a lightweight contextual selector for state-dependent policy selection during online adaptation. Each policy in the online candidate set is treated as an arm, while the current environment state serves as the context. Formally, we denote the candidate policy set by $\{\pi^{k}\}_{k=1}^K$ and write the LinUCB statistics compactly as $B(A,b)$, where $A=\{A_k\}_{k=1}^K$ and $b=\{b_k\}_{k=1}^K$, with each $A_k\in\mathbb{R}^{d\times d}$ initialized to $I_d$ and each $b_k\in\mathbb{R}^d$ initialized to $0_d$. At each online interaction step $t$, LinUCB computes a confidence-adjusted score for each candidate policy from the current state $s_t$:
\begin{align}
     &u_t^{k} \gets (A_k^{-1}b_k)^Ts_t+\omega\sqrt{s_t^T A_k^{-1} s_t}, k=1,...,K
     \notag
     \\&P_t \leftarrow \mathrm{Softmax}(u_t^1,\ldots,u_t^K),
     \notag
     \\&k_t \leftarrow \arg\max_k P_t(k).
     \label{eq:appendx_1}
\end{align}
Here $P_t$ denotes the policy-preference vector obtained by softmax-normalizing the LinUCB scores. Softmax preserves the ranking of arms, so selecting $\arg\max_k P_t(k)$ executes the policy with the highest raw LinUCB score. Figures~\ref{fig:effectiveness} and~\ref{fig:effectiveness_appendix} visualize $P_t$ throughout online adaptation. After executing the selected policy, HiP3 observes the immediate environment reward $r_t$ and updates only the statistics of the selected arm:
\begin{align}
 &A_{k_t}\leftarrow A_{k_t}+\eta_B s_ts_t^T,
 \notag
 \\&b_{k_t}\leftarrow b_{k_t}+\eta_B r_t s_t,
 \label{eq:appendx_2}
\end{align}
where $k_t=\arg\max_k P_t(k)$ and $\eta_B=0.05$ is the update rate used in our implementation to stabilize the online adaptation of the bandit statistics.

We use the observed environment reward as direct target-task feedback, avoiding additional model- or critic-based value estimation, while long-horizon policy improvement is handled separately by the SAC updates during online fine-tuning. The selector is applied throughout the 100K-step online fine-tuning stage used in our experiments. A comparison with a nonlinear KernelUCB selector is reported in Appendix~\ref{app:linucb_analysis}, showing that increasing selector capacity does not improve performance under the same online protocol.

\section{Experimental Setup}\label{app:exp_setup}

\subsection{MuJoCo Simulation Environment}\label{app:environments}
We conduct the experiments on continuous control tasks using the MuJoCo~\citep{todorov2012mujoco} simulation environment. For offline pre-training, we use the standard D4RL Gym benchmark~\citep{fu2020d4rl}, which provides datasets with diverse qualities. Our HiP3 and all the baseline methods are tested on three tasks: Walker2d, Hopper, and HalfCheetah. For each task, these algorithms are evaluated on datasets with four different quality levels: random-v2, medium-v2, medium-replay-v2, and medium-expert-v2. These datasets are generated as follows. "Random" is generated by collecting 1 million steps of rollouts from a randomly initialized policy. "Medium" is generated by collecting 1 million steps of rollouts from a partially trained SAC. "Medium-replay" is generated by recording all samples in the replay buffer observed during training until the policy reaches a medium level of performance. "Medium-expert" is generated by mixing 1 million steps of expert-level data with 1 million steps of sub-optimal data. 

In the online stage, we categorize the experiments into two types of online tasks: the same online tasks and novel online tasks. For the former, the same offline task environments are used for the online tasks, while for the latter we introduce changes in both dynamics and reward structure, following common practices for inducing distribution shifts in continuous control benchmarks. Specifically, we modify the environments as follows:
\begin{itemize}
    \item Gravity increased from $9.8$ $\rightarrow$ $11.5$
    \item Friction coefficients scaled by $1.2\times$
    \item Forward-reward weight scaled by $0.85$
    \item Control-cost weight increased to $0.2$
    \item Noise scales are set to $0.2$, $0.1$, and $0.05$ for HalfCheetah, Hopper, and Walker2d, respectively.
\end{itemize}
These modifications introduce noticeable distribution shifts and increase task difficulty compared to the standard settings.


\subsection{Implementation Details}\label{appendix: baseline implementation}
We first describe the baseline implementations. Cal-QL~\citep{nakamoto2023cal}, AWAC~\citep{nair2020awac}, IQL~\citep{kostrikov2021offline} and CQL~\citep{kumar2020conservative} are evaluated in the offline-to-online settings provided by CORL~\citep{tarasov2023corl}, using the official CORL implementations. For MBPO~\citep{janner2019trust}, PEX~\citep{zhang2023policy}, WSRL~\citep{zhouefficient2025}, SO2~\citep{zhang2024perspective}, FQL~\citep{fql_park2025} and EDIS~\citep{liuenergy2024}, we use the official implementations released by their authors. When conducting the comparison for PEX, we employed the same O2O setting provided by the official code. For MBPO, we learned the policies from scratch with the hyper-parameters provided by the official implementation. For FamO2O~\citep{wang2023train}, we have referenced their published results, which involved running 1M steps for online fine-tuning.

All O2O methods except FamO2O are pretrained for 1M offline steps and then fine-tuned for 100K online steps. MBPO is trained from scratch for 100K steps. For HiP3, candidate initialization evaluates each offline policy with one target-task rollout and selects the top-$K$ candidates, using a rollout horizon of at most 500 steps per policy. The ranking trajectories are used only for candidate initialization and are not added to the online replay buffer or used for policy or model updates. The resulting candidate set is then fine-tuned for 100K online steps following the procedure described in Section~\ref{sec: method}. The reported 100K-step budget counts only environment transitions used for online fine-tuning. The one-time rollout-based candidate ranking uses additional target-environment interactions outside this fine-tuning budget.

\subsection{Computing Infrastructure}\label{app:comput_infra}
Given the large number of experiments, our experiments were conducted using a mix of GPUs, including NVIDIA RTX-3090, NVIDIA V100, NVIDIA A100, and NVIDIA A6000. NVIDIA RTX-3090 was run on Ubuntu operating system, while the other GPUs used CentOS. The software dependencies, implementation details, and instructions for reproducing our experiments are provided in the README of our public code repository.

\subsection{Computational Cost}\label{app:comput_cost}

HiP3 constructs multiple policies in the offline stage, but this does not multiply the total offline optimization budget by the number of policies. All three matched variants use the same total 1M-step offline policy-optimization budget and 100K-step online fine-tuning budget. For HiP3 and HiP3-LS, the offline budget is distributed across the policy pool rather than assigned independently to each policy. This matched protocol allows us to separate differences in wall-clock cost from differences in the total amount of policy optimization.

Table~\ref{tbl:matched_computation_cost} reports the end-to-end wall-clock cost of the three variants under the same model-based backbone, rollout configuration, hardware, and optimization budgets. The single-policy variant requires 5.8h for offline training and 8.4h for online fine-tuning, for a total of 14.2h. HiP3-LS increases the offline time to 6.3h while keeping the online time unchanged at 8.4h. Since the total offline optimization budget is fixed, this increase reflects the additional computation associated with maintaining and optimizing multiple candidate policies rather than additional policy-optimization steps. Full HiP3 requires 7.1h offline and 8.5h online, for a total of 15.6h. Relative to HiP3-LS, the additional offline computation is associated with MOSAIC: HiP3-LS uses fixed linear scalarization, whereas MOSAIC additionally evaluates reference alignment and combines multiple objective gradients through MGDA during optimization.

MOSAIC further expands the policy pool through warm-started neighborhood search rather than training every additional candidate from scratch. After optimizing a policy for a reference vector, HiP3 reuses the resulting policy as the initialization for nearby reference vectors. This allows additional trade-off regions to be populated without linearly increasing the optimization budget with the number of candidate policies. The total offline policy-optimization budget therefore remains fixed at 1M steps.

At the online stage, HiP3 fine-tunes only the policy selected at the current interaction step rather than updating all candidate policies. The policy pool therefore does not multiply the online policy-optimization cost. LinUCB is computationally lightweight, maintaining a matrix $A \in \mathbb{R}^{n \times d \times d}$ and a vector $b \in \mathbb{R}^{n \times d}$, where $n < 10$ is the number of candidate policies and $d < 50$ is the observation dimension. Consistent with this design, the online wall-clock times of the three matched variants are nearly identical in Table~\ref{tbl:matched_computation_cost}. The reported online time for HiP3 also includes the one-time rollout ranking used to initialize the candidate set, which takes less than 10 minutes in our MuJoCo experiments on the same NVIDIA A6000 setup.

\begin{table}[H]
\begin{center}
\begin{tabular}{lccc}
\toprule
\textbf{Variant} & \textbf{Offline Time (h)} & \textbf{Online Time (h)} & \textbf{End-to-End Time (h)}\\
\midrule
Single-policy  & 5.8h  & 8.4h  & 14.2h \\
HiP3-LS  & 6.3h  & 8.4h  & 14.7h  \\
HiP3     & 7.1h  & 8.5h  & 15.6h  \\
\bottomrule
\end{tabular}\vspace{3pt}
\caption{End-to-end computational cost of matched HiP3 variants. All variants use the same total 1M-step offline policy-optimization budget and 100K-step online fine-tuning budget. Wall-clock times are averaged over 3 seeds on a common NVIDIA A6000 GPU.}
\label{tbl:matched_computation_cost}
\end{center}
\end{table}

For additional runtime context, Tables~\ref{tbl:offline_computation_cost} and~\ref{tbl:online_computation_cost} report stage-wise wall-clock times for representative baselines evaluated on the same NVIDIA A6000 platform. These measurements provide absolute resource references across different algorithmic implementations, while Table~\ref{tbl:matched_computation_cost} provides the controlled comparison used to isolate the computational cost introduced by HiP3 itself.

\begin{table}[htb]
\begin{center}
\begin{tabular}{lcccc}
\toprule
\textbf{Method} & \textbf{HiP3 (offline)} & \textbf{Cal-QL (offline)} & \textbf{PEX (offline)} & \textbf{MOPO (offline)} \\
\midrule
Hopper       & 7.5h  & 9.2h  & 2.9h & 5.9h  \\
HalfCheetah  & 8.3h  & 8.7h  & 4.0h & 7.0h  \\
Walker2d     & 5.6h  & 7.5h  & 2.7h & 5.2h  \\
\bottomrule
\end{tabular}\vspace{3pt}
\caption{Computation cost comparison for offline stage}
\label{tbl:offline_computation_cost}
\end{center}
\end{table}

\begin{table}[htb]
\begin{center}
\begin{tabular}{lcc}
\toprule
\textbf{Method} & \textbf{HiP3 (online)} & \textbf{MBPO (online)} \\
\midrule
Hopper       & 7.7h  & 7.8h  \\
HalfCheetah  & 8.5h  & 8.9h  \\
Walker2d     & 9.4h  & 9.7h  \\
\bottomrule
\end{tabular}\vspace{3pt}
\caption{Computation cost comparison for online stage}
\label{tbl:online_computation_cost}
\end{center}
\end{table}

\subsection{The Hyper-parameters of HiP3}\label{app:hyper_hip3}

For the training of HiP3 on the MuJoCo tasks, the hyper-parameters for the offline and the online stages can be found in Table~\ref{tbl:mujoco_hyperparameters}(a) and Table~\ref{tbl:mujoco_hyperparameters}(b), respectively.

\begin{table}[htb]
\centering
\footnotesize 
\setlength{\tabcolsep}{4pt} 

\begin{subtable}[t]{0.48\textwidth}
    \centering
    \begin{tabular}[t]{lp{2.6cm}} 
    \toprule
    \textbf{Hyper-parameter} & \textbf{Value}  \\
    \midrule
    Number of hidden units & $256$ \\
    Number of hidden layers & $2$ \\
    Optimizer & Adam\\
    Rollout length       & Walker2d: 1 \\
                         & HalfCheetah: 3 \\
                         & Hopper: 5 \\
    Rollout batch size & 50\\
    Updates per step & $1$ \\
    Autotune SAC $\alpha$ & True \\
    SAC $\alpha$ learning rate  & $3\times10^{-4}$ \\
    Max SAC $\alpha$ & 1.0\\
    Uncertainty-squashing constant $\kappa$ & 1.5\\
    Learning rate  & $3\times10^{-4}$ \\
    Discount factor & $0.99$ \\
    Reference-vector similarity threshold $\psi$ & $0.999$ \\
    Target update rate & $5\times10^{-3}$ \\
    Batch size & 256\\
    Number of reference vectors & $5$ \\
    Environment steps & $5\times(180K+2\times10K)$\\
    $(\tau_a,\tau_b)$ of reference vectors & $(0.1, 0.9)$ \\
    Number of offline policies & $100$ \\
    \bottomrule
    \end{tabular}
    \caption{Hyper-parameters for offline RL}
    \label{tbl:offline_hyper-parameters}
\end{subtable}
\hfill
\begin{subtable}[t]{0.48\textwidth}
    \centering
    \begin{tabular}[t]{lp{2.6cm}} 
    \toprule
    \textbf{Hyper-parameter} & \textbf{Value}  \\
    \midrule
    Number of hidden units & $256$ \\
    Number of hidden layers & $2$ \\
    Optimizer & Adam\\
    Rollout length  & $1$\\    
    Rollout batch size & $5\times10^4$\\
    Model training frequency & 500\\
    Model retain epochs & 5\\
    Updates per step  & 15 \\  
    Autotune SAC $\alpha$ & True\\
    SAC $\alpha$ learning rate  & $3\times10^{-4}$ \\
    Max SAC $\alpha$ & 1.0\\
    Learning rate  & $3\times10^{-4}$ \\
    Discount factor & $0.99$ \\
    Target update rate & $5\times10^{-3}$ \\
    Batch size & 256\\
    Number of epochs & 100\\
    Environment steps per epoch & 1000\\
    Number of online policies & $5$ \\
    \bottomrule
    \end{tabular}
    \caption{Hyper-parameters for online RL}
    \label{tbl:online_hyper-parameters}
\end{subtable}
\caption{The hyper-parameters of HiP3 used for offline and online learning on the MuJoCo tasks.}
\label{tbl:mujoco_hyperparameters}
\end{table}

\section{Additional Empirical Results}\label{app:add_emp_result}
\subsection{Generality of Optimism-Pessimism Crossover}\label{app:motivation}
Figure~\ref{fig:pareto_front} provides an illustrative example showing that optimistic policies can adapt better to novel online tasks, while pessimistic policies can perform better when the online task remains close to the offline distribution. To examine whether this phenomenon is specific to a single setting, we conduct a controlled study across four representative offline datasets.

We consider two MuJoCo domains (HalfCheetah and Hopper) and two offline dataset qualities (random and medium-replay). For each offline policy pool, we identify an optimistic and a pessimistic policy according to their optimism and pessimism objectives, respectively. The two policies are selected before observing the online task and are then independently fine-tuned for 100K online steps using the same online optimization procedure. No Top-K ranking, LinUCB selection, or policy switching is applied in this analysis.

\begin{table}[H]
\centering
\footnotesize
\setlength{\tabcolsep}{5pt}
\begin{tabular}{lcccc}
\toprule
\textbf{Offline Dataset} 
& \multicolumn{2}{c}{\textbf{Same Online Task}}
& \multicolumn{2}{c}{\textbf{Novel Online Task}}\\
\cmidrule(lr){2-3}
\cmidrule(lr){4-5}
& \textbf{Optimistic} & \textbf{Pessimistic}
& \textbf{Optimistic} & \textbf{Pessimistic}\\
\midrule

halfcheetah-r 
& 48.4 & \textbf{56.1}
& \textbf{57.6} & 12.8\\

halfcheetah-mr 
& 56.2 & \textbf{85.9}
& \textbf{54.5} & 53.0\\

hopper-r 
& 45.7 & \textbf{100.6}
& \textbf{74.2} & 31.6\\

hopper-mr 
& 78.6 & \textbf{102.2}
& \textbf{86.5} & 37.7\\

\midrule
\textbf{mean}
& 57.2 & \textbf{86.2}
& \textbf{68.2} & 33.8\\

\bottomrule
\end{tabular}
\vspace{4pt}
\caption{
Generality of the optimism-pessimism crossover behavior.
Optimistic and pessimistic policies are selected before observing the online task and independently fine-tuned for 100K online steps without Top-$K$ ranking, LinUCB selection, or policy switching. The results show a consistent crossover between adaptation on the same and novel online tasks.
}
\label{tab:crossover}
\end{table}

Across all four settings, the relative ordering consistently reverses after introducing the online distribution shift: pessimistic policies perform better on the same online tasks, while optimistic policies achieve higher performance on the novel online tasks. These results suggest that the preferred optimism-pessimism trade-off depends on the online task rather than being fixed during offline training.

\FloatBarrier
\subsection{Results on Additional Domains}\label{app:add_domain}
We further evaluate HiP3 on additional domains beyond standard D4RL locomotion tasks, including high-dimensional manipulation (Adroit) and navigation (Maze2D). As shown in Table~\ref{tab:generalization}, HiP3 achieves the highest mean across the two Adroit tasks and the highest Maze2D score among the compared O2O baselines. These results extend the empirical evaluation to manipulation and navigation settings.
\begin{table}[H]
\centering
\footnotesize
\setlength{\tabcolsep}{5pt}
\begin{tabular}{llcccc}
\toprule
\textbf{Domain} & \textbf{Task} & \textbf{HiP3} & \textbf{Cal-QL} & \textbf{AWAC} & \textbf{CQL} \\
\midrule
\multirow{3}{*}{Adroit} 
& pen-human-v1   & 35.8±10.2 & 27.6±24.8 & \textbf{36.6±10.6} & 1.2±5.0\\
& pen-cloned-v1  & \textbf{54.6±4.7} & -3.2±0.1 & 47.3±11.7 & 13.2±6.4\\
& \textbf{mean}  & \textbf{45.2±7.5} & 12.2±12.5 & 42.0±11.2 & 7.2±5.7\\
\midrule
Maze2D 
& medium-dense-v1 & \textbf{142.4±21.2} & 77.3±21.2 & 48.7±26.4 & 59.2±24.1 \\
\bottomrule
\end{tabular}
\vspace{4pt}
\caption{Generalization to additional domains. We evaluate HiP3 on high-dimensional manipulation tasks (Adroit) and navigation tasks (Maze2D), and compare with widely-used O2O RL baselines (Cal-QL, AWAC, CQL). HiP3 achieves competitive or superior performance across domains.}
\label{tab:generalization}
\end{table}

\FloatBarrier
\subsection{Transient Performance Degradation and Recovery Analysis}\label{appendix:performance_drop}
We further analyze the transient degradation observed during early online fine-tuning. This analysis complements the relative performance-drop comparison in Table~\ref{tbl:performance_drop_comparison} by measuring the absolute minimum performance and recovery behavior.

\begin{table}[H]
\centering
\footnotesize
\setlength{\tabcolsep}{4pt}

\begin{subtable}[t]{0.49\textwidth}
\centering
\begin{tabular}{lccc}
\toprule
\textbf{Method} & \textbf{HiP3} & \textbf{MOPO+MBPO} & \textbf{Cal-QL} \\
\midrule
halfcheetah-r-v2 & 8.3 & 18.7 & 17.2\\
halfcheetah-mr-v2 & 7.6 & 15.0 & 11.6\\
hopper-r-v2 & 14.6 & 2.8 & 3.4\\
hopper-mr-v2 & 20.4 & 28.5 & 21.5\\
\midrule
mean & 12.7 & 16.3 & 13.4\\
\bottomrule
\end{tabular}
\caption{Novel online tasks}
\label{tbl:performance_drop_novel}
\end{subtable}
\hfill
\begin{subtable}[t]{0.49\textwidth}
\centering
\begin{tabular}{lccc}
\toprule
\textbf{Method} & \textbf{HiP3} & \textbf{MOPO+MBPO} & \textbf{Cal-QL} \\
\midrule
halfcheetah-r-v2 & 1.2 & 3.6 & 2.1\\
halfcheetah-mr-v2 & 1.4 & 9.8 & 4.8\\
hopper-r-v2 & 12.1 & 1.5 & 4.2\\
hopper-mr-v2 & 16.7 & 23.1 & 2.6\\
\midrule
mean & 7.9 & 9.5 & 3.4\\
\bottomrule
\end{tabular}
\caption{Same online tasks}
\label{tbl:performance_drop_same}
\end{subtable}

\caption{Relative performance drop comparison on same and novel online tasks.
The drop is measured between the pre-fine-tuning performance and the minimum performance during online adaptation.
(Notations: r = random, mr = medium-replay.)}
\label{tbl:performance_drop_comparison}
\end{table}

For each run, we define the minimum performance as the lowest normalized score observed during online fine-tuning. The recovery time is defined as the first checkpoint after degradation where the performance returns to the pre-fine-tuning level.

\begin{table}[H]
\centering
\footnotesize
\setlength{\tabcolsep}{6pt}
\begin{tabular}{lcc}
\toprule
\textbf{Task} & 
\textbf{Minimum Normalized Score} &
\textbf{Recovery Step, Median [Range]} \\
\midrule

Same-HC-r 
& $33.2 \pm 1.3$ 
& $8K$ [$7K$, $9K$] \\

Same-HC-mr 
& $46.6 \pm 15.7$ 
& $11K$ [$7K$, $18K$] \\

Same-Hop-r 
& $14.5 \pm 1.0$ 
& $24K$ [$15K$, $31K$] \\

Same-Hop-mr 
& $43.0 \pm 5.9$ 
& $18K$ [$12K$, $28K$] \\

\midrule

Novel-HC-r 
& $27.0 \pm 2.8$ 
& $7K$ [$6K$, $10K$] \\

Novel-HC-mr 
& $39.3 \pm 21.2$ 
& $9K$ [$7K$, $12K$] \\

Novel-Hop-r 
& $11.2 \pm 1.8$ 
& $26K$ [$13K$, $38K$] \\

Novel-Hop-mr 
& $27.7 \pm 8.6$ 
& $17K$ [$10K$, $24K$] \\

\bottomrule
\end{tabular}
\vspace{4pt}
\caption{
Absolute transient degradation and recovery analysis during online fine-tuning. Minimum normalized scores are averaged over three seeds. Recovery steps are reported as median [minimum, maximum] across seeds.
}
\label{tab:performance_recovery}
\end{table}

Across the 24 runs, the median recovery time is 12K online steps: 18/24 runs recover within 20K steps, and all runs recover within 38K steps. The degradation is generally smaller on HalfCheetah tasks, whereas Hopper tasks exhibit larger and longer transient drops.

These results suggest that although HiP3 achieves strong final adaptation performance, practical deployment may require safeguards during early fine-tuning, such as retaining the offline policy as a fallback or delaying aggressive online updates.

\FloatBarrier
\subsection{Hyperparameter Sensitivity Analysis}\label{app:hyper_sensitive}
We further evaluate the sensitivity of HiP3 to its method-specific hyperparameters, including the uncertainty-squashing constant $\kappa$, reference-vector range $(\tau_a,\tau_b)$, perturbation size $\epsilon$, and similarity threshold $\psi$.

All experiments use the same protocol as the main experiments, and only one hyperparameter is changed at a time while keeping all other configurations fixed. We evaluate representative novel online tasks across different domains and report the mean normalized return. The default configurations are shared across all evaluation tasks without task-specific adjustment.

\begin{table}[!htbp]
\centering
\footnotesize
\setlength{\tabcolsep}{4pt}
\begin{tabular}{llccccc}
\toprule
\textbf{Sweep} & \textbf{Value} &
\textbf{HC-r} &
\textbf{HC-mr} &
\textbf{Hop-r} &
\textbf{Hop-mr} &
\textbf{Mean}\\
\midrule

$\kappa$ & 0.5
&66.8$\pm$3.0
&66.4$\pm$3.9
&92.2$\pm$5.1
&94.6$\pm$8.3
&80.0\\

$\kappa$ & \textbf{1.5}
&\textbf{63.5$\pm$2.2}
&\textbf{69.4$\pm$3.6}
&\textbf{95.2$\pm$5.5}
&\textbf{100.1$\pm$11.0}
&\textbf{82.1}\\

$\kappa$ & 2.5
&61.9$\pm$0.4
&68.7$\pm$1.6
&96.7$\pm$7.3
&97.4$\pm$10.6
&81.2\\

\midrule

$(\tau_a,\tau_b)$ & (0.05,0.95)
&64.1$\pm$2.6
&69.3$\pm$2.0
&88.0$\pm$8.2
&97.6$\pm$17.5
&79.8\\

$(\tau_a,\tau_b)$ & \textbf{(0.1,0.9)}
&\textbf{63.5$\pm$2.2}
&\textbf{69.4$\pm$3.6}
&\textbf{95.2$\pm$5.5}
&\textbf{100.1$\pm$11.0}
&\textbf{82.1}\\

$(\tau_a,\tau_b)$ & (0.2,0.8)
&63.9$\pm$2.3
&70.2$\pm$3.1
&89.5$\pm$3.5
&95.8$\pm$16.7
&79.9\\

\bottomrule
\end{tabular}
\vspace{6pt}
\caption{
Sensitivity analysis of the uncertainty-squashing constant $\kappa$ and reference-vector range $(\tau_a,\tau_b)$. Default configurations are bolded. Each non-default setting changes only the corresponding hyperparameter while keeping all other configurations fixed.
}
\label{tab:sensitivity_kappa_ref}
\end{table}

\begin{table}[!htbp]
\centering
\footnotesize
\setlength{\tabcolsep}{4pt}
\begin{tabular}{llccccc}
\toprule
\textbf{Sweep} & \textbf{Value} &
\textbf{HC-r} &
\textbf{HC-mr} &
\textbf{Hop-r} &
\textbf{Hop-mr} &
\textbf{Mean}\\
\midrule

$\epsilon$ & 0.025
&62.7$\pm$1.5
&67.3$\pm$3.6
&91.4$\pm$5.1
&96.0$\pm$4.3
&79.4\\

$\epsilon$ & \textbf{0.05}
&\textbf{63.5$\pm$2.2}
&\textbf{69.4$\pm$3.6}
&\textbf{95.2$\pm$5.5}
&\textbf{100.1$\pm$11.0}
&\textbf{82.1}\\

$\epsilon$ & 0.075
&58.7$\pm$3.6
&72.3$\pm$2.5
&98.5$\pm$5.0
&94.2$\pm$7.9
&80.9\\

\midrule

$\psi$ & 0.995
&64.1$\pm$4.8
&72.5$\pm$3.3
&94.3$\pm$7.2
&94.7$\pm$12.1
&81.4\\

$\psi$ & \textbf{0.999}
&\textbf{63.5$\pm$2.2}
&\textbf{69.4$\pm$3.6}
&\textbf{95.2$\pm$5.5}
&\textbf{100.1$\pm$11.0}
&\textbf{82.1}\\

$\psi$ & 0.9995
&63.0$\pm$2.7
&68.7$\pm$4.8
&90.6$\pm$4.6
&104.5$\pm$6.4
&81.7\\

\bottomrule
\end{tabular}
\vspace{6pt}
\caption{
Sensitivity analysis of the perturbation size $\epsilon$ and similarity threshold $\psi$. Default configurations are bolded. The same experimental protocol is used across all settings.
}
\label{tab:sensitivity_eps_psi}
\end{table}

Across all tested hyperparameters, the maximum variation in mean normalized return is 2.7 points, and no task-specific adjustment is performed.

\FloatBarrier
\subsection{Complete Benchmark Results and Additional Analyses}\label{app:full_results}
This section presents additional empirical analyses in six parts: model and training sensitivity, online policy-selection analysis, complete learning curves on the same and novel online tasks, additional ablations on policy-pool construction and selection, complete benchmark tables, and extended baseline comparisons.

\subsubsection{Model and Training Sensitivity}\label{app:addition_ablation}
We first examine the sensitivity of HiP3 to world-model training. Table~\ref{tbl:model_impact} varies the hyper-parameter max-model-epoch-since-update controlling model update epochs ($\mathrm{mmepoch}\in\{1,5,10\}$). The default setting ($\mathrm{mmepoch}=5$) performs comparably to longer training while clearly outperforming the under-trained model.

\begin{figure}[H]
    \centering
    \includegraphics[width=0.85\linewidth]{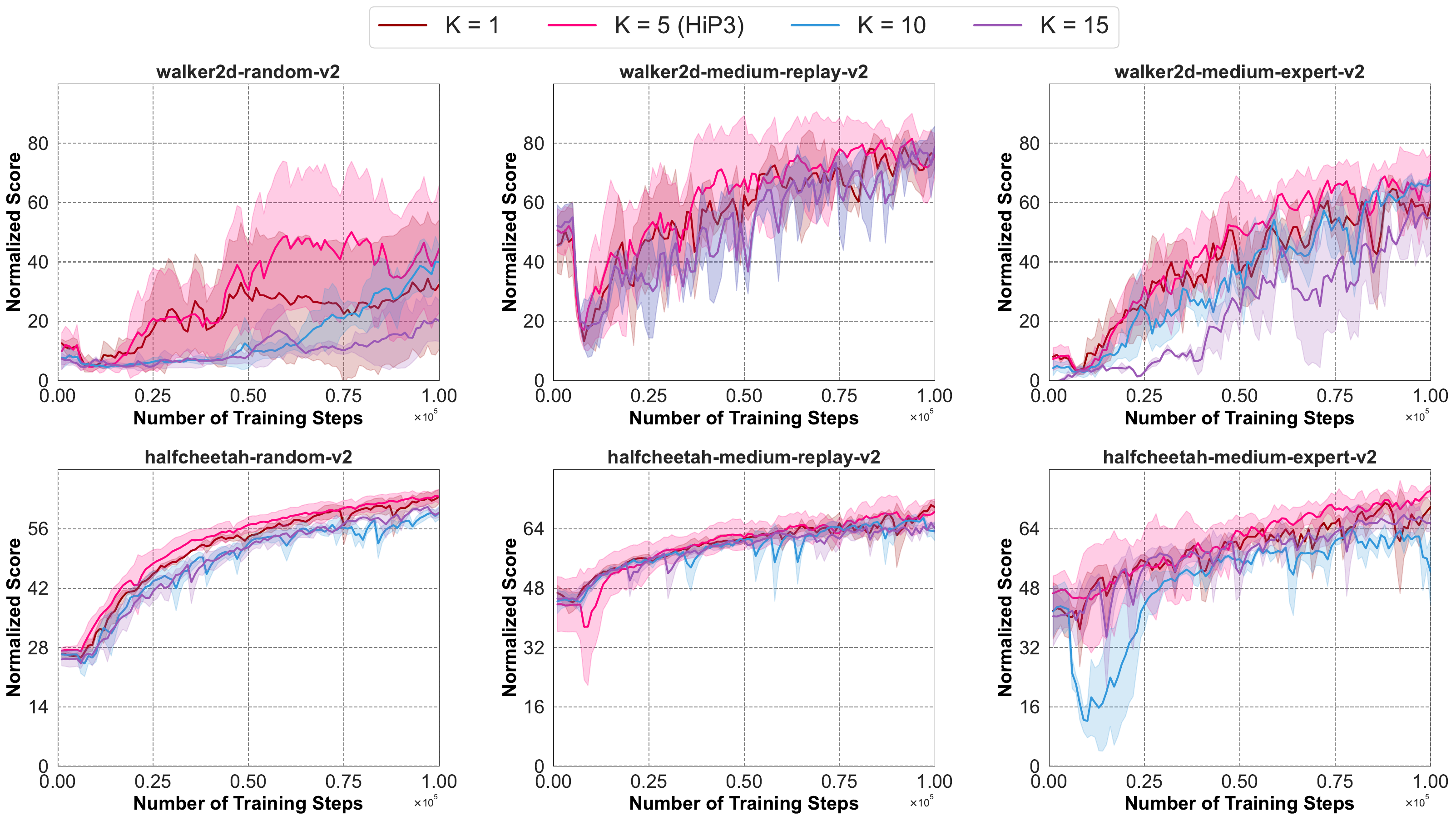}
    \caption{Ablation study on the number of policies $K$ when fine-tuning the policy pool on six novel online tasks. The normalized score is averaged over 5 random seeds (each with 10 evaluations). The curves exclusively illustrate the performance during the online stage, where LinUCB is used for policy selection from the pool.}
    \label{fig:ablation_Policy_num_appendix}
\end{figure}

Additionally, Figure~\ref{fig:ablation_Policy_num_appendix} studies the number of candidate policies $K$. Too many candidates receive insufficient online updates under the limited interaction budget, whereas too few candidates reduce policy diversity. We therefore use $K=5$, which performs best on most evaluated tasks.

\begin{table}[H]
\centering
\footnotesize
\setlength{\tabcolsep}{5pt}
\begin{tabular}{lccc}
\toprule
\textbf{Method} & \textbf{HiP3 (default, mmepoch=5)} & \textbf{HiP3 (mmepoch=1)} & \textbf{HiP3 (mmepoch=10)} \\
\midrule
halfcheetah-random-v2       & 63.5±2.2  & 48.9±3.7 & 63.1±2.6  \\
halfcheetah-medium-replay-v2  & 69.4±3.6  & 56.7±3.1 & 66.3±2.3 \\
hopper-random-v2       & 95.2±5.5  & 25.0±9.8 & 92.3±9.0  \\
hopper-medium-replay-v2  & 100.1±11.0  & 26.1±8.2 & 100.8±6.9 \\
\midrule
mean & 82.1±5.5 & 39.2±6.2 & 80.6±5.2\\
\bottomrule
\end{tabular}\vspace{5pt}
\caption{Ablation study on world model training impact}
\label{tbl:model_impact}
\end{table}



\FloatBarrier
\subsubsection{Online Policy-Selection Analysis}
\label{app:linucb_analysis}
We first examine whether state-dependent policy selection benefits from a more expressive nonlinear selector. Table~\ref{tbl:bandit_impact} compares the default linear LinUCB with a nonlinear KernelUCB variant based on random Fourier features. Under the same online adaptation protocol and bandit feedback, LinUCB achieves a mean score of 82.1 compared with 77.8 for KernelUCB, indicating that increasing the selector's function-class capacity does not improve performance in the evaluated settings. We then analyze how the resulting LinUCB preferences evolve during online adaptation. Figure~\ref{fig:effectiveness} visualizes the policy preferences $P_t$ obtained by softmax-normalizing the LinUCB scores and used for state-dependent policy selection. To summarize executed selections over the full online run, Table~\ref{tab:linucb_frequency} reports the dominant-policy selection frequency,
\[
\max_k \frac{1}{T}\sum_{t=1}^{T}\mathbf{1}[k_t=k],
\]
defined as the fraction of online steps assigned to the most frequently selected candidate.

\begin{table}[H]
\caption{Impact of bandit model capacity on online adaptation performance}
\label{tbl:bandit_impact}
\centering
\footnotesize
\setlength{\tabcolsep}{5pt}
\begin{tabular}{lcc}
\toprule
\textbf{Method} & \textbf{HiP3 (LinUCB)} & \textbf{HiP3 (KernelUCB)}\\
\midrule
halfcheetah-random-v2       & 63.5±2.2 & 62.0±2.8\\
halfcheetah-medium-replay-v2  & 69.4±3.6  & 63.5±4.9\\
hopper-random-v2       & 95.2±5.5  & 86.8±11.7\\
hopper-medium-replay-v2  & 100.1±11.0  & 99.0±2.8 \\
\midrule
mean & 82.1±5.5 & 77.8±5.6\\
\bottomrule
\end{tabular}\vspace{5pt}
\end{table}



\FloatBarrier
\subsubsection{Complete Learning Curves on Same and Novel Online Tasks}
\begin{figure}[H]
    \centering
    \includegraphics[width=0.8\linewidth]{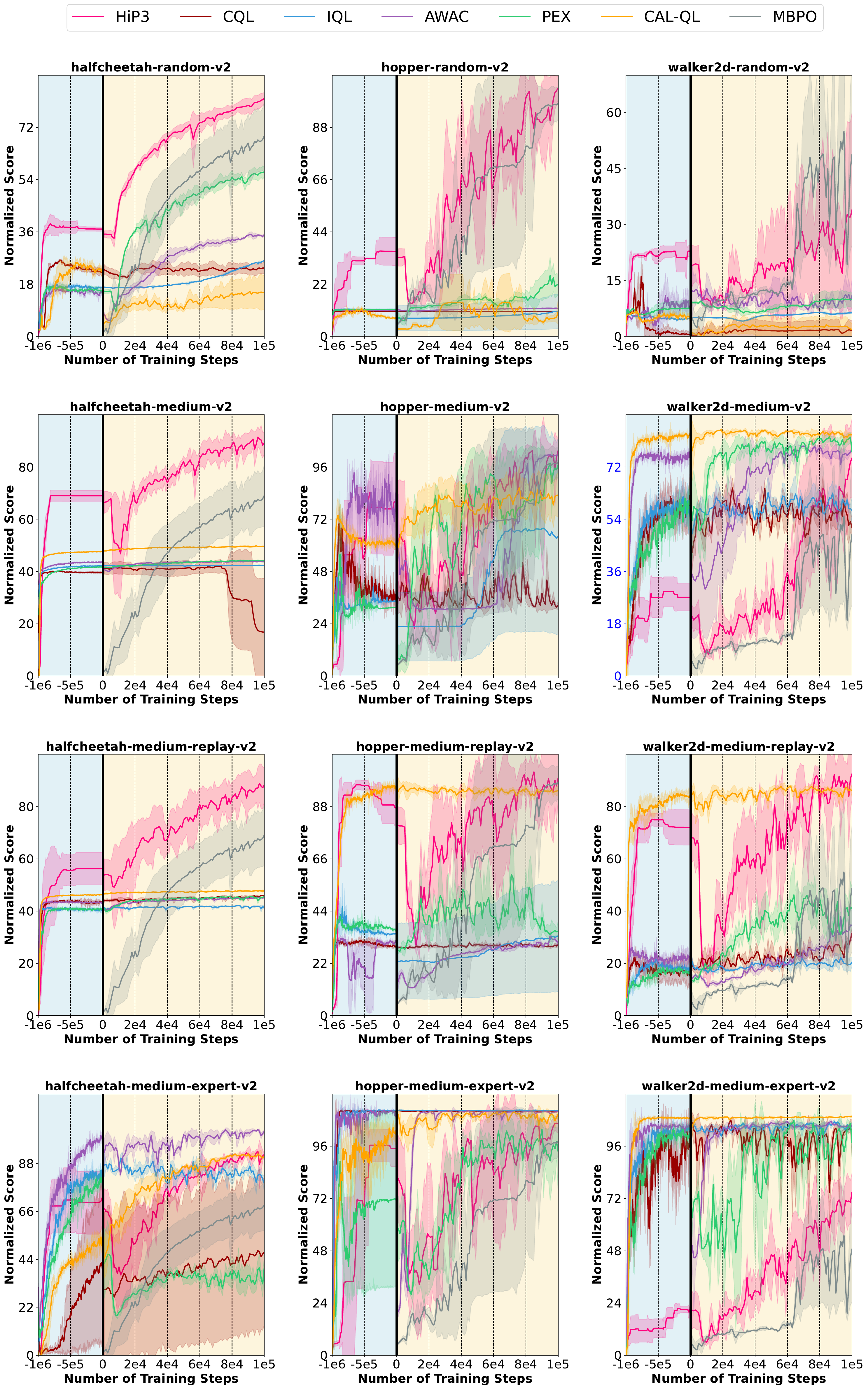}
    \caption{Offline-to-online learning curves on \textbf{the same online tasks}. Curves show mean over five seeds; shaded regions denote standard deviation.}
    \label{fig:comparision_same_env}
\end{figure}

\begin{figure}[H]
    \centering
    \includegraphics[width=0.8\linewidth]{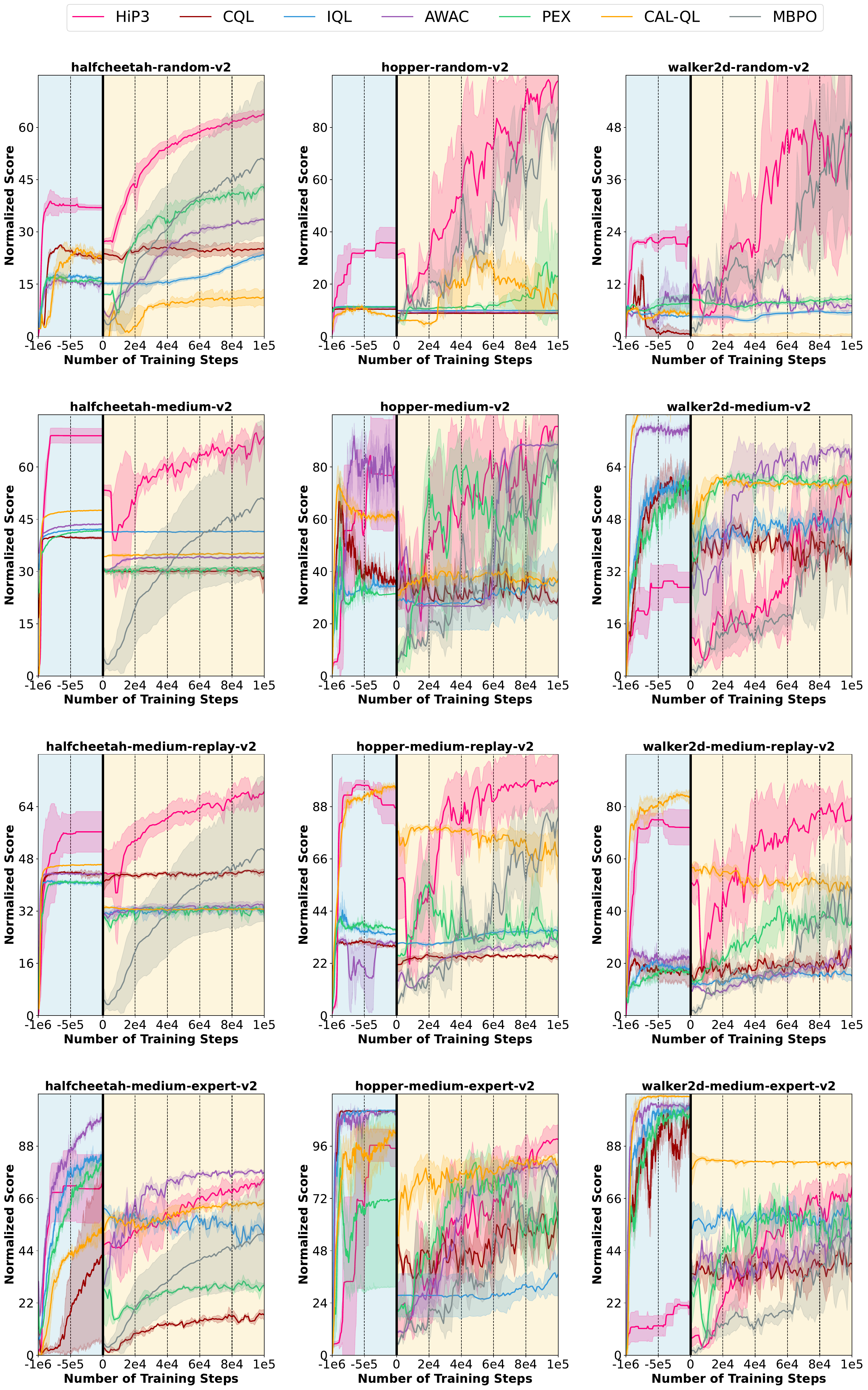}\vspace{-5pt}
    \caption{Offline-to-online learning curves on \textbf{the novel online tasks}. These tasks are generated via modifying the gravity, noise scales, and reward functions of the original MuJoCo tasks. Curves show mean over five seeds; shaded regions denote standard deviation.}
    \label{fig:comparision_diff_env}
\end{figure}

We provide the full offline-to-online learning curves on the same and novel online tasks in Figures~\ref{fig:comparision_same_env} and~\ref{fig:comparision_diff_env}, respectively. These figures complement the representative plots reported in the main paper and show that the overall trend in favor of HiP3 is consistent across datasets.

\subsubsection{Additional Ablations on Policy Pool and Policy Selection}
\begin{figure}[H]
    \centering
    \includegraphics[width=0.85\linewidth]{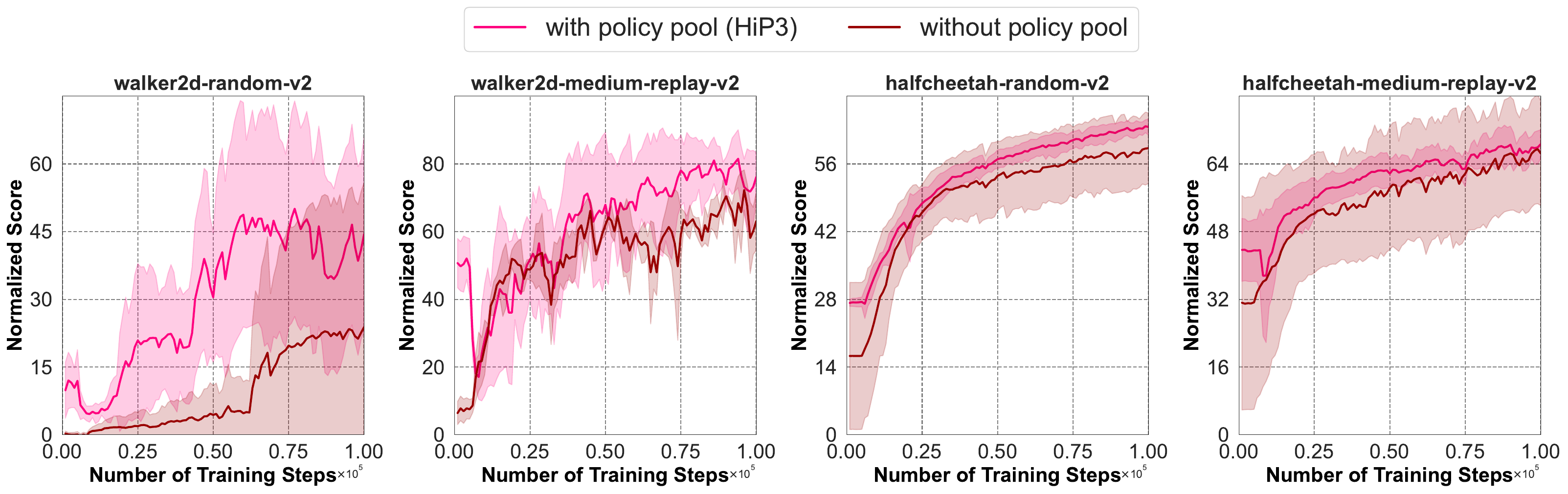}
    \caption{Complete comparison between HiP3 and the single-policy variant on four representative novel online tasks. Curves show mean and standard deviation over five seeds.}
    \label{fig:ablation_policypool_appendix}
\end{figure}

\begin{figure}[H]
    \centering
    \includegraphics[width=0.85\linewidth]{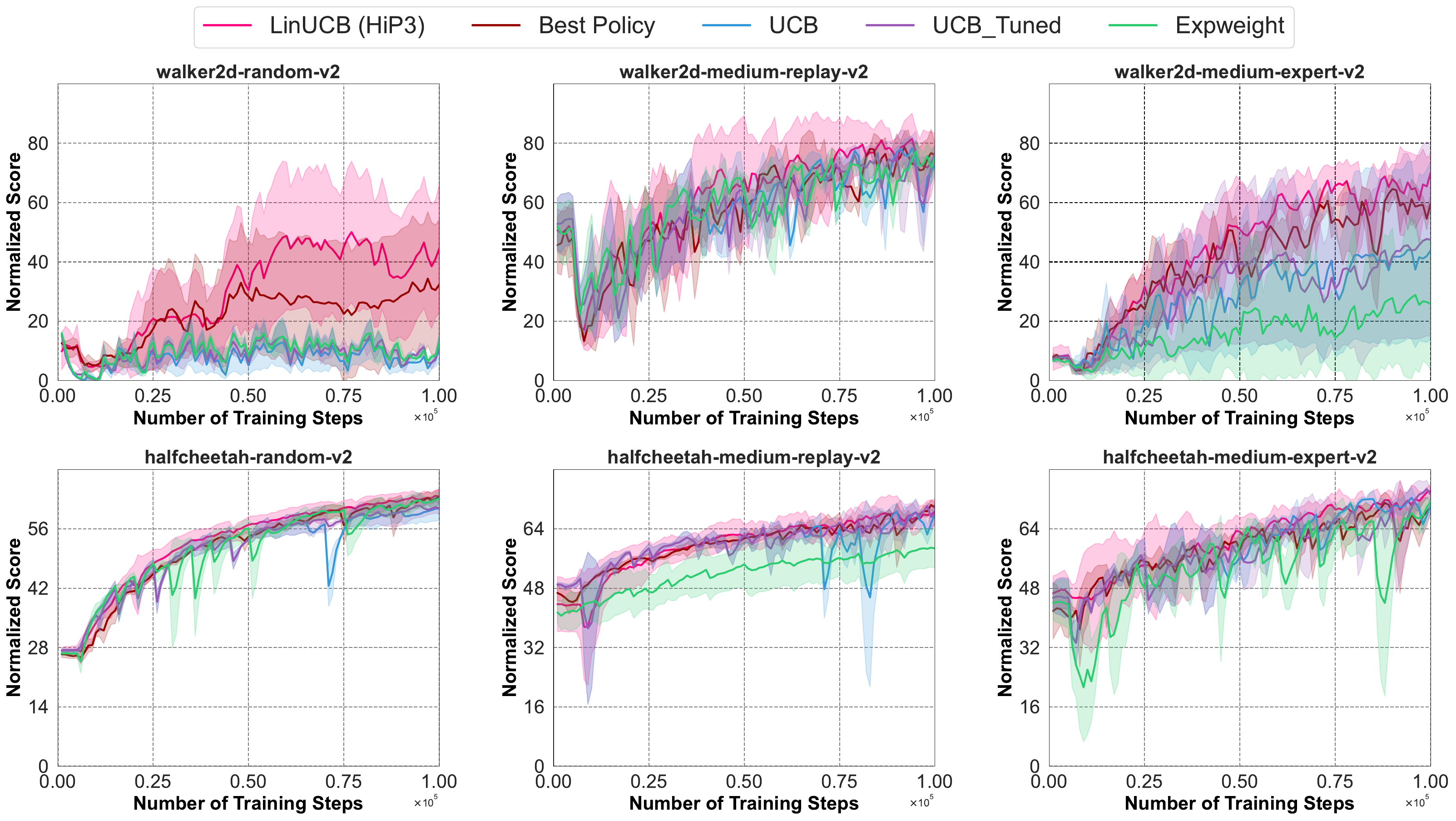}
    \caption{Ablation study on the policy selection algorithms for six novel online tasks. The normalized score is averaged over 5 random seeds (each with 10 evaluations). The curves exclusively illustrate the performance during the online stage, where various bandit algorithms are evaluated.}
    \label{fig:ablation_select_algorithm_appendix}
\end{figure}

We next report the complete ablations on policy-pool construction and policy selection.  Figure~\ref{fig:ablation_policypool_appendix} studies the effect of removing the policy pool, while Figure~\ref{fig:ablation_select_algorithm_appendix} compares different policy-selection algorithms. Table~\ref{tbl:ablation_ref_num} evaluates the number of reference vectors under a fixed offline optimization budget, and Table~\ref{tab:single_policy_ablation} extends the single-policy comparison to all 12 novel online tasks.


\begin{table}[H]
\begin{center}
\begin{tabular}{lccc}
\toprule
\textbf{Reference number} & \textbf{ref\_vec=5 (HiP3)} & \textbf{ref\_vec=4} & \textbf{ref\_vec=2} \\
\midrule
halfcheetah-r-v2 & 63.5±2.2 & 63.3±1.2 & 63.5±1.5\\
halfcheetah-mr-v2 & 69.4±3.6 & 67.7±4.5 & 68.4±3.8\\
hopper-r-v2 & 95.2±5.5 & 86.1±8.2 & 87.5±10.4\\
hopper-mr-v2 & 100.1±11.0 & 90.4±10.7 & 93.8±12.3\\
\midrule
mean & 82.1±5.5 & 76.8±6.2 & 78.3±7.0\\
\bottomrule
\end{tabular}\vspace{3pt}
\caption{Ablation study for different number of reference vectors. For fairness, we fix the total training budget to 1M steps and allocate it evenly across reference vectors. When too many reference vectors are used, each policy receives insufficient training, leading to performance deterioration. Conversely, too few reference vectors reduce the diversity of the policy pool. The ablation study shows that using five reference vectors achieves the best balance between policy diversity and overall performance.}
\label{tbl:ablation_ref_num}
\end{center}
\end{table}

\begin{table}[H]
\centering
\footnotesize
\setlength{\tabcolsep}{5pt}
\begin{tabular}{lcc}
\toprule
\textbf{Novel Online Task} 
& \textbf{HiP3} 
& \textbf{Single-Policy HiP3} \\
\midrule

halfcheetah-r-v2
& $63.5 \pm 2.2$
& $57.1 \pm 11.8$ \\

halfcheetah-medium-v2
& $69.3 \pm 2.1$
& $54.2 \pm 8.3$ \\

halfcheetah-medium-replay-v2
& $69.4 \pm 3.6$
& $67.9 \pm 15.4$ \\

halfcheetah-medium-expert-v2
& $74.7 \pm 2.0$
& $70.3 \pm 4.9$ \\

\midrule

hopper-r-v2
& $95.2 \pm 5.5$
& $83.2 \pm 0.3$ \\

hopper-medium-v2
& $95.4 \pm 3.8$
& $88.5 \pm 1.0$ \\

hopper-medium-replay-v2
& $100.1 \pm 11.0$
& $86.5 \pm 3.0$ \\

hopper-medium-expert-v2
& $99.1 \pm 7.6$
& $84.0 \pm 2.3$ \\

\midrule

walker2d-r-v2
& $49.2 \pm 26.7$
& $23.1 \pm 28.6$ \\

walker2d-medium-v2
& $59.2 \pm 6.4$
& $47.7 \pm 36.4$ \\

walker2d-medium-replay-v2
& $77.3 \pm 10.4$
& $61.0 \pm 9.7$ \\

walker2d-medium-expert-v2
& $74.6 \pm 3.9$
& $43.9 \pm 14.6$ \\

\midrule

\textbf{Mean}
& $\mathbf{77.3 \pm 7.1}$
& $64.0 \pm 12.9$ \\

\bottomrule
\end{tabular}
\vspace{6pt}
\caption{
Complete comparison between HiP3 and its single-policy variant on all 12 novel online tasks.
The single-policy variant removes the offline Pareto policy pool while retaining the same model-based online fine-tuning procedure.
HiP3 uses the five seeds reported in the main results, while the single-policy variant uses three seeds per task.
Both methods use the same offline data, environment model, SAC-based backbone, and online fine-tuning budget.
}
\label{tab:single_policy_ablation}
\end{table}


\FloatBarrier
\clearpage
\subsubsection{Complete Benchmark Tables}
Table~\ref{table:baseline_comparison_mujoco_appendix} reports the complete per-task comparison between HiP3 and the main baselines on the same and novel online tasks. These detailed tables complement the summary table in the main paper and make the performance differences across datasets explicit.

\begin{table}[H]
\setlength{\tabcolsep}{2pt}
\small
\centering
\begin{subtable}[t]{1.0\textwidth}
\resizebox{\linewidth}{!}{
\begin{tabular}{c@{\hspace{5pt}}lr@{\hspace{-1pt}}lr@{\hspace{-1pt}}lr@{\hspace{-1pt}}lr@{\hspace{-1pt}}lr@{\hspace{-1pt}}lr@{\hspace{-1pt}}lr@{\hspace{-1pt}}lr@{\hspace{-1pt}}l}
\toprule
\multicolumn{1}{l}{} &  & \multicolumn{2}{c}{CQL} & \multicolumn{2}{c}{IQL} & \multicolumn{2}{c}{FamO2O*} & \multicolumn{2}{c}{AWAC} & \multicolumn{2}{c}{PEX} & \multicolumn{2}{c}{Cal-QL} & \multicolumn{2}{c}{MBPO} & \multicolumn{2}{c}{HiP3}  \\
\midrule
\multirow{3}{*}{\rotatebox[origin=c]{90}{Random}} & HalfCheetah &\colorbox{white}{23.7}&{\color[HTML]{525252}$\pm$2.1}  &\colorbox{white}{26.4}&{\color[HTML]{525252}$\pm$0.1} &\colorbox{white}{}&{$-$} &\colorbox{white}{35.6}&{\color[HTML]{525252}$\pm$0.5}  &\colorbox{white}{56.9}&{\color[HTML]{525252}$\pm$1.2}  &\colorbox{white}{15.4}&{\color[HTML]{525252}$\pm$5.6} &\colorbox{white}{66.0}&{\color[HTML]{525252}$\pm$15.4} &\colorbox{mine}{82.6}&{\color[HTML]{525252}$\pm$2.3} \\
                                  & Hopper &\colorbox{white}{10.5}&{\color[HTML]{525252}$\pm$0.1}  &\colorbox{white}{10.4}&{\color[HTML]{525252}$\pm$7.3}  &\colorbox{white}{}&{$-$} &\colorbox{white}{11.7}&{\color[HTML]{525252}$\pm$0.2}  &\colorbox{white}{22.2}&{\color[HTML]{525252}$\pm$4.1}  &\colorbox{white}{12.7}&{\color[HTML]{525252}$\pm$8.5} &\colorbox{white}{98.2}&{\color[HTML]{525252}$\pm$6.1} &\colorbox{mine}{105.7}&{\color[HTML]{525252}$\pm$4.1} \\
          						  & Walker2d &\colorbox{white}{1.7}&{\color[HTML]{525252}$\pm$2.7}  &\colorbox{white}{6.3}&{\color[HTML]{525252}$\pm$0.2}  &\colorbox{white}{}&{$-$} &\colorbox{white}{12.2}&{\color[HTML]{525252}$\pm$6.7}  &\colorbox{white}{9.8}&{\color[HTML]{525252}$\pm$2.2}  &\colorbox{white}{1.6}&{\color[HTML]{525252}$\pm$1.3} &\colorbox{mine}{51.4}&{\color[HTML]{525252}$\pm$35.9} &\colorbox{white}{35.7}&{\color[HTML]{525252}$\pm$26.0} \\
\multirow{3}{*}{\rotatebox[origin=c]{90}{Medium}} & HalfCheetah &\colorbox{white}{16.7}&{\color[HTML]{525252}$\pm$20.7}  &\colorbox{white}{42.4}&{\color[HTML]{525252}$\pm$0.4}  &{}&\colorbox{white}{53.4} &\colorbox{white}{43.6}&{\color[HTML]{525252}$\pm$0.8}  &\colorbox{white}{44.1}&{\color[HTML]{525252}$\pm$0.3}  &\colorbox{white}{49.7}&{\color[HTML]{525252}$\pm$0.3} &\colorbox{white}{66.0}&{\color[HTML]{525252}$\pm$15.4} &\colorbox{mine}{90.1}&{\color[HTML]{525252}$\pm$4.1} \\
                                  & Hopper &\colorbox{white}{33.6}&{\color[HTML]{525252}$\pm$3.2}  &\colorbox{white}{65.4}&{\color[HTML]{525252}$\pm$45.7} &{}&\colorbox{white}{82.8} &\colorbox{white}{101.5}&{\color[HTML]{525252}$\pm$0.4}  &\colorbox{white}{99.4}&{\color[HTML]{525252}$\pm$1.2}  &\colorbox{white}{83.3}&{\color[HTML]{525252}$\pm$12.3} &\colorbox{white}{98.2}&{\color[HTML]{525252}$\pm$6.1}&\colorbox{mine}{106.7}&{\color[HTML]{525252}$\pm$3.0} \\
          						  & Walker2d &\colorbox{white}{49.1}&{\color[HTML]{525252}$\pm$8.2}  &\colorbox{white}{56.0}&{\color[HTML]{525252}$\pm$10.2} &{}&\colorbox{white}{82.8} &\colorbox{white}{79.0}&{\color[HTML]{525252}$\pm$1.7}  &\colorbox{white}{72.0}&{\color[HTML]{525252}$\pm$1.2} &\colorbox{mine}{83.1}&{\color[HTML]{525252}$\pm$2.2} &\colorbox{white}{51.4}&{\color[HTML]{525252}$\pm$35.9} &\colorbox{white}{76.4}&{\color[HTML]{525252}$\pm$16.0} \\
\multirow{3}{*}{\rotatebox[origin=c]{90}{\shortstack{Medium\\-replay}}} & HalfCheetah &\colorbox{white}{46.0}&{\color[HTML]{525252}$\pm$0.1}  &\colorbox{white}{41.5}&{\color[HTML]{525252}$\pm$0.5}  &{}&\colorbox{white}{51.0}  &\colorbox{white}{45.3}&{\color[HTML]{525252}$\pm$0.1}  &\colorbox{white}{45.9}&{\color[HTML]{525252}$\pm$0.1} &\colorbox{white}{47.8}&{\color[HTML]{525252}$\pm$0.1}  &\colorbox{white}{66.0}&{\color[HTML]{525252}$\pm$15.4} &\colorbox{mine}{89.6}&{\color[HTML]{525252}$\pm$2.7} \\
                                  & Hopper &\colorbox{white}{30.5}&{\color[HTML]{525252}$\pm$0.9}  &\colorbox{white}{33.9}&{\color[HTML]{525252}$\pm$23.6}  &{}&\colorbox{white}{92.2} &\colorbox{white}{29.6}&{\color[HTML]{525252}$\pm$2.8}  &\colorbox{white}{36.0}&{\color[HTML]{525252}$\pm$7.0} &\colorbox{white}{95.1}&{\color[HTML]{525252}$\pm$1.4} &\colorbox{white}{98.2}&{\color[HTML]{525252}$\pm$6.1} &\colorbox{mine}{103.4}&{\color[HTML]{525252}$\pm$3.7} \\
								  & Walker2d &\colorbox{white}{33.8}&{\color[HTML]{525252}$\pm$3.6}  &\colorbox{white}{21.3}&{\color[HTML]{525252}$\pm$1.0} &{}&\colorbox{white}{88.6} &\colorbox{white}{34.5}&{\color[HTML]{525252}$\pm$8.4}  &\colorbox{white}{40.6}&{\color[HTML]{525252}$\pm$7.3}  &\colorbox{white}{85.5}&{\color[HTML]{525252}$\pm$4.9} &\colorbox{white}{51.4}&{\color[HTML]{525252}$\pm$35.9} &\colorbox{mine}{94.1}&{\color[HTML]{525252}$\pm$12.1} \\
\multirow{3}{*}{\rotatebox[origin=c]{90}{\shortstack{Medium\\-expert}}} & HalfCheetah &\colorbox{white}{48.4}&{\color[HTML]{525252}$\pm$31.8}  &\colorbox{white}{76.7}&{\color[HTML]{525252}$\pm$6.8} &{}&\colorbox{white}{91.8} &\colorbox{mine}{103.9}&{\color[HTML]{525252}$\pm$1.5}  &\colorbox{white}{48.3}&{\color[HTML]{525252}$\pm$5.6}  &\colorbox{white}{92.5}&{\color[HTML]{525252}$\pm$1.6} &\colorbox{white}{66.0}&{\color[HTML]{525252}$\pm$15.4} &\colorbox{white}{92.8}&{\color[HTML]{525252}$\pm$3.4} \\
                                  & Hopper &\colorbox{white}{111.4}&{\color[HTML]{525252}$\pm$0.6}  &\colorbox{mine}{112.4}&{\color[HTML]{525252}$\pm$0.1} &{}&\colorbox{white}{90.1} &\colorbox{white}{112.3}&{\color[HTML]{525252}$\pm$0.4}  &\colorbox{white}{89.0}&{\color[HTML]{525252}$\pm$31.6} &\colorbox{white}{112.2}&{\color[HTML]{525252}$\pm$0.2} &\colorbox{white}{98.2}&{\color[HTML]{525252}$\pm$6.1} &\colorbox{white}{106.5}&{\color[HTML]{525252}$\pm$4.7} \\
    								  & Walker2d &\colorbox{white}{93.5}&{\color[HTML]{525252}$\pm$14.9}  &\colorbox{white}{105.1}&{\color[HTML]{525252}$\pm$2.9} &{}&\colorbox{white}{110.6} &\colorbox{white}{105.8}&{\color[HTML]{525252}$\pm$1.8}  &\colorbox{white}{89.0}&{\color[HTML]{525252}$\pm$18.0}  &\colorbox{mine}{109.6}&{\color[HTML]{525252}$\pm$0.7} &\colorbox{white}{51.4}&{\color[HTML]{525252}$\pm$35.9} &\colorbox{white}{79.3}&{\color[HTML]{525252}$\pm$10.6} \\
\midrule
\multicolumn{1}{l}{Total}              & Mean &\colorbox{white}{41.6}&{\color[HTML]{525252}$\pm$7.4} &\colorbox{white}{49.8}&{\color[HTML]{525252}$\pm$8.2} &{}&\colorbox{white}{82.6} &\colorbox{white}{59.6}&{\color[HTML]{525252}$\pm$2.2} &\colorbox{white}{54.4}&{\color[HTML]{525252}$\pm$6.6} &\colorbox{white}{65.7}&{\color[HTML]{525252}$\pm$3.3} &\colorbox{white}{71.9}&{\color[HTML]{525252}$\pm$19.1} &\colorbox{mine}{88.6}&{\color[HTML]{525252}$\pm$7.7} \\
\bottomrule
\end{tabular}}
\subcaption{Fine-tuning performance on the same online tasks}
\end{subtable}
\vspace{8pt}

\begin{subtable}[t]{1.0\textwidth}
\resizebox{\linewidth}{!}{
\begin{tabular}{c@{\hspace{5pt}}lr@{\hspace{-1pt}}lr@{\hspace{-1pt}}lr@{\hspace{-1pt}}lr@{\hspace{-1pt}}lr@{\hspace{-1pt}}lr@{\hspace{-1pt}}lr@{\hspace{-1pt}}lr@{\hspace{-1pt}}l}
\toprule
\multicolumn{1}{l}{} &  & \multicolumn{2}{c}{CQL} & \multicolumn{2}{c}{IQL} & \multicolumn{2}{c}{AWAC} & \multicolumn{2}{c}{PEX} & \multicolumn{2}{c}{Cal-QL} & \multicolumn{2}{c}{MBPO} & \multicolumn{2}{c}{HiP3}  \\
\midrule
\multirow{3}{*}{\rotatebox[origin=c]{90}{Random}} & HalfCheetah &\colorbox{white}{24.7}&{\color[HTML]{525252}$\pm$1.1}  &\colorbox{white}{23.1}&{\color[HTML]{525252}$\pm$1.3} &\colorbox{white}{33.2}&{\color[HTML]{525252}$\pm$0.4}  &\colorbox{white}{42.2}&{\color[HTML]{525252}$\pm$2.6} &\colorbox{white}{11.1}&{\color[HTML]{525252}$\pm$2.5} &\colorbox{white}{47.1}&{\color[HTML]{525252}$\pm$22.1} &\colorbox{mine}{63.5}&{\color[HTML]{525252}$\pm$2.2} \\
                                  & Hopper &\colorbox{white}{8.9}&{\color[HTML]{525252}$\pm$0.1}  &\colorbox{white}{9.9}&{\color[HTML]{525252}$\pm$0.0} &\colorbox{white}{9.2}&{\color[HTML]{525252}$\pm$0.1}  &\colorbox{white}{23.5}&{\color[HTML]{525252}$\pm$17.2} &\colorbox{white}{11.7}&{\color[HTML]{525252}$\pm$4.7} &\colorbox{white}{86.4}&{\color[HTML]{525252}$\pm$4.6} &\colorbox{mine}{95.2}&{\color[HTML]{525252}$\pm$5.5} \\
          						  & Walker2d &\colorbox{white}{-0.8}&{\color[HTML]{525252}$\pm$0.4}  &\colorbox{white}{5.5}&{\color[HTML]{525252}$\pm$0.6} &\colorbox{white}{7.0}&{\color[HTML]{525252}$\pm$1.3}  &\colorbox{white}{8.5}&{\color[HTML]{525252}$\pm$0.7}  &\colorbox{white}{-0.6}&{\color[HTML]{525252}$\pm$1.1} &\colorbox{white}{44.6}&{\color[HTML]{525252}$\pm$11.4} &\colorbox{mine}{49.2}&{\color[HTML]{525252}$\pm$26.7} \\
\multirow{3}{*}{\rotatebox[origin=c]{90}{Medium}} & HalfCheetah &\colorbox{white}{29.3}&{\color[HTML]{525252}$\pm$2.5}  &\colorbox{white}{41.5}&{\color[HTML]{525252}$\pm$0.2} &\colorbox{white}{33.9}&{\color[HTML]{525252}$\pm$0.4}  &\colorbox{white}{30.4}&{\color[HTML]{525252}$\pm$0.9} &\colorbox{white}{35.4}&{\color[HTML]{525252}$\pm$0.1} &\colorbox{white}{47.1}&{\color[HTML]{525252}$\pm$22.1} &\colorbox{mine}{69.3}&{\color[HTML]{525252}$\pm$2.1} \\
                                  & Hopper &\colorbox{white}{31.5}&{\color[HTML]{525252}$\pm$5.4}  &\colorbox{white}{38.4}&{\color[HTML]{525252}$\pm$12.9} &\colorbox{white}{88.5}&{\color[HTML]{525252}$\pm$0.1}  &\colorbox{white}{74.8}&{\color[HTML]{525252}$\pm$15.9} &\colorbox{white}{37.6}&{\color[HTML]{525252}$\pm$4.6} &\colorbox{white}{86.4}&{\color[HTML]{525252}$\pm$4.6}  &\colorbox{mine}{95.4}&{\color[HTML]{525252}$\pm$3.8} \\
          						  & Walker2d &\colorbox{white}{43.1}&{\color[HTML]{525252}$\pm$9.7}  &\colorbox{white}{46.9}&{\color[HTML]{525252}$\pm$1.7} &\colorbox{mine}{70.4}&{\color[HTML]{525252}$\pm$0.3}  &\colorbox{white}{61.5}&{\color[HTML]{525252}$\pm$1.0} &\colorbox{white}{59.6}&{\color[HTML]{525252}$\pm$0.4} &\colorbox{white}{44.6}&{\color[HTML]{525252}$\pm$11.4} &\colorbox{white}{59.2}&{\color[HTML]{525252}$\pm$6.4} \\
\multirow{3}{*}{\rotatebox[origin=c]{90}{\shortstack{Medium\\-replay}}} & HalfCheetah &\colorbox{white}{44.5}&{\color[HTML]{525252}$\pm$0.1}  &\colorbox{white}{32.3}&{\color[HTML]{525252}$\pm$2.1} &\colorbox{white}{29.9}&{\color[HTML]{525252}$\pm$3.8}  &\colorbox{white}{30.4}&{\color[HTML]{525252}$\pm$3.5} &\colorbox{white}{32.6}&{\color[HTML]{525252}$\pm$0.2} &\colorbox{white}{47.1}&{\color[HTML]{525252}$\pm$22.1} &\colorbox{mine}{69.4}&{\color[HTML]{525252}$\pm$3.6} \\
                                  & Hopper &\colorbox{white}{24.7}&{\color[HTML]{525252}$\pm$1.5}  &\colorbox{white}{35.2}&{\color[HTML]{525252}$\pm$2.0} &\colorbox{white}{29.6}&{\color[HTML]{525252}$\pm$2.8}  &\colorbox{white}{39.4}&{\color[HTML]{525252}$\pm$14.7} &\colorbox{white}{65.9}&{\color[HTML]{525252}$\pm$14.0} &\colorbox{white}{86.4}&{\color[HTML]{525252}$\pm$4.6}  &\colorbox{mine}{100.1}&{\color[HTML]{525252}$\pm$11.0} \\
								  & Walker2d &\colorbox{white}{21.1}&{\color[HTML]{525252}$\pm$0.7}  &\colorbox{white}{14.5}&{\color[HTML]{525252}$\pm$1.3} &\colorbox{white}{20.0}&{\color[HTML]{525252}$\pm$3.2}  &\colorbox{white}{38.0}&{\color[HTML]{525252}$\pm$4.3} &\colorbox{white}{46.4}&{\color[HTML]{525252}$\pm$3.7} &\colorbox{white}{44.6}&{\color[HTML]{525252}$\pm$11.4} &\colorbox{mine}{77.3}&{\color[HTML]{525252}$\pm$10.4} \\
\multirow{3}{*}{\rotatebox[origin=c]{90}{\shortstack{Medium\\-expert}}} & HalfCheetah &\colorbox{white}{16.6}&{\color[HTML]{525252}$\pm$2.9}  &\colorbox{white}{59.3}&{\color[HTML]{525252}$\pm$5.3} &\colorbox{mine}{77.6}&{\color[HTML]{525252}$\pm$0.7}  &\colorbox{white}{29.9}&{\color[HTML]{525252}$\pm$1.0} &\colorbox{white}{64.4}&{\color[HTML]{525252}$\pm$1.9}  &\colorbox{white}{47.1}&{\color[HTML]{525252}$\pm$22.1}  &\colorbox{white}{74.7}&{\color[HTML]{525252}$\pm$2.0} \\
                                  & Hopper &\colorbox{white}{68.7}&{\color[HTML]{525252}$\pm$6.8}  &\colorbox{white}{38.7}&{\color[HTML]{525252}$\pm$11.8} &\colorbox{white}{88.8}&{\color[HTML]{525252}$\pm$4.4}  &\colorbox{white}{50.9}&{\color[HTML]{525252}$\pm$9.5} &\colorbox{white}{88.0}&{\color[HTML]{525252}$\pm$5.5} &\colorbox{white}{86.4}&{\color[HTML]{525252}$\pm$4.6}  &\colorbox{mine}{99.1}&{\color[HTML]{525252}$\pm$7.6} \\
    								  & Walker2d &\colorbox{white}{38.2}&{\color[HTML]{525252}$\pm$9.8}  &\colorbox{white}{58.8}&{\color[HTML]{525252}$\pm$11.4} &\colorbox{white}{50.6}&{\color[HTML]{525252}$\pm$8.5}  &\colorbox{white}{54.3}&{\color[HTML]{525252}$\pm$2.8} &\colorbox{mine}{79.2}&{\color[HTML]{525252}$\pm$3.4} &\colorbox{white}{44.6}&{\color[HTML]{525252}$\pm$11.4}&\colorbox{white}{74.6}&{\color[HTML]{525252}$\pm$3.9} \\
\midrule
\multicolumn{1}{l}{Total}              & Mean &\colorbox{white}{29.2}&{\color[HTML]{525252}$\pm$3.4} &\colorbox{white}{33.7}&{\color[HTML]{525252}$\pm$4.2} &\colorbox{white}{44.9}&{\color[HTML]{525252}$\pm$4.9} &\colorbox{white}{40.3}&{\color[HTML]{525252}$\pm$6.2} &\colorbox{white}{44.3}&{\color[HTML]{525252}$\pm$3.6} &\colorbox{white}{59.4}&{\color[HTML]{525252}$\pm$12.7} &\colorbox{mine}{77.3}&{\color[HTML]{525252}$\pm$7.1} \\
\bottomrule
\end{tabular}}
\subcaption{Fine-tuning performance on novel online tasks}
\end{subtable}
\caption{Performance comparison on the same and novel online tasks. The novel online tasks are generated by adjusting the gravity, noise scales, and reward functions of the original MuJoCo environments. $*$ indicates the results reported by the paper. Note that FamO2O reports results with a 1M-step online fine-tuning budget and is therefore not directly budget-matched to the 100K-step setting used for the other methods. We fine-tune all other methods for 100K steps, using the same hyper-parameter settings provided by the original papers. Scores for each dataset are normalized according to the D4RL benchmark~\citep{fu2020d4rl}: $\textbf{Normalized Score} = 100\times\frac{\text{return}-\text{random return}}{\text{expert return}-\text{random return}}$.}
\label{table:baseline_comparison_mujoco_appendix}
\end{table}

\FloatBarrier
\subsubsection{Extended Baseline Comparisons}
Finally, we include additional comparisons with recent model-based reinforcement learning baselines that are relevant to HiP3. Table~\ref{table:fowm} compares HiP3 with FOWM~\citep{feng2023finetuning} on the same online tasks, while Table~\ref{table:tdmpc2} compares HiP3 with TD-MPC2 on novel online tasks using a 100K-step online training horizon.

As shown in Table~\ref{table:fowm}, HiP3 outperforms FOWM on the Hopper tasks, even though FOWM is fine-tuned for 500K online environment steps while HiP3 uses only 100K. For TD-MPC2~\citep{hansen2024tdmpc2}, we evaluate all runs for 100K online training steps, following the limited-interaction setting used in our main evaluation. We set the model size to 5, following the recommendation in the TD-MPC2 paper, and use the default values for the remaining hyper-parameters. Table~\ref{table:tdmpc2} shows that HiP3 substantially outperforms TD-MPC2 under this limited-budget setting.

\begin{table}[H]
\centering
\begin{subtable}[t]{0.48\linewidth}
\centering
\begin{tabular}{lrr}
\toprule
{Task}                            & {HiP3}   & {FOWM$^*$} \\
\midrule
{hopper-medium-v2}                & {\textbf{106.7}}  & {100.7}    \\
{hopper-medium-replay-v2 }        & {\textbf{103.4}}  & {93.5}     \\
\midrule
{mean}                            & {\textbf{105.1}} & {97.1} \\
\bottomrule
\end{tabular}
\subcaption{{Same online tasks with FOWM}}
\label{table:fowm}
\vspace{8pt}
\end{subtable}
\hfill
\begin{subtable}[t]{0.48\linewidth}
\centering
\begin{tabular}{lrr}
\toprule
{Task}                & {HiP3}   & {TD-MPC2} \\
\midrule
{halfcheetah}         &{\textbf{69.2}}   & {37.3}    \\
{hopper}              & {\textbf{97.5}}   & {3.1}     \\
{walker2d}            & {\textbf{65.1}}   & {2.8}     \\
\midrule
{mean}               & {\textbf{77.3}} & {14.4} \\
\bottomrule
\end{tabular}
\subcaption{{Novel online tasks with TD-MPC2}}
\label{table:tdmpc2}
\end{subtable}
\caption{Comparisons with model-based reinforcement learning baselines. * indicates results reported by the paper. FOWM results are reported from its paper; TD-MPC2 is evaluated for 100K online training steps.}
\end{table}







\end{document}